\documentclass[10pt,twocolumn,letterpaper]{article}

\usepackage[pagenumbers]{cvpr} 

\usepackage{graphicx}
\usepackage{amsmath}
\usepackage{amssymb}
\usepackage{booktabs}
\usepackage{multicol}
\usepackage{multirow}
\usepackage{arydshln}
\usepackage{comment}

\usepackage[pagebackref,breaklinks,colorlinks]{hyperref}

\newcommand{\zl}[1]{\textcolor{blue}{ZL: #1}}

\newcommand{\toh}[1]{\textcolor{magenta}{TH: #1}}
\newcommand*{\affaddr}[1]{#1} 
\newcommand*{\affmark}[1][*]{\textsuperscript{#1}}
\newcommand*{\email}[1]{\texttt{#1}}

\usepackage[capitalize]{cleveref}
\crefname{section}{Sec.}{Secs.}
\Crefname{section}{Section}{Sections}
\Crefname{table}{Table}{Tables}
\crefname{table}{Tab.}{Tabs.}

\def\confName{CVPR}
\def\confYear{2023}

\begin{document}

\title{SmartBrush: Text and Shape Guided Object Inpainting with Diffusion Model}

\author{%
Shaoan Xie \affmark[1]\footnotemark , Zhifei Zhang\affmark[2], Zhe Lin\affmark[2], Tobias Hinz\affmark[2], and Kun Zhang\affmark[1,3]\\
\affaddr{\affmark[1]Carnegie Mellon University}\\
\affaddr{\affmark[2]Adobe Research}\\
\affaddr{\affmark[3]Mohamed bin Zayed University of Artificial Intelligence}\\
\email{shaoan@cmu.edu, \{zzhang, zlin, thinz\}@adobe.com, kunz1@cmu.edu}\\
}

\twocolumn[{%
\renewcommand\twocolumn[1][]{#1}%

\maketitle

\setlength{\tabcolsep}{2pt}
\def\qualheight{2.2cm}
\begin{center}
\def\fsh{\small}
\begin{tabular}{cccccccc}
   & \fsh{Input+Mask} & \fsh{Stable Diffusion} & \fsh{Blended Diffusion}  & \fsh{GLIDE} & \fsh{Stable Inpainting} & \fsh{DALLE-2} & \fsh{SmartBrush} \\
        \raisebox{3em}{\rotatebox[origin=c]{90}{\fsh{flamingo}}} &
        \frame{\includegraphics[height=\qualheight]{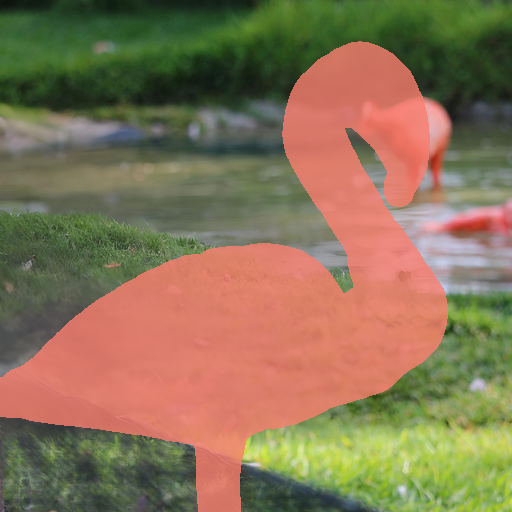}}& \frame{\includegraphics[height=\qualheight]{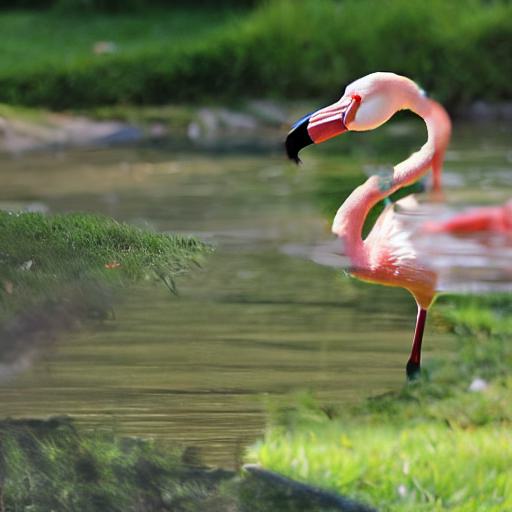}}&
        \frame{\includegraphics[height=\qualheight]{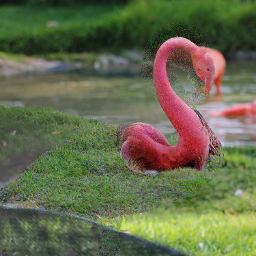}}&
        \frame{\includegraphics[height=\qualheight]{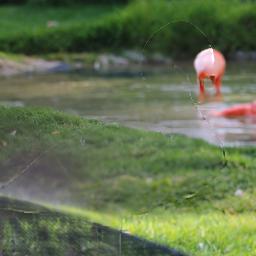}}& \frame{\includegraphics[height=\qualheight]{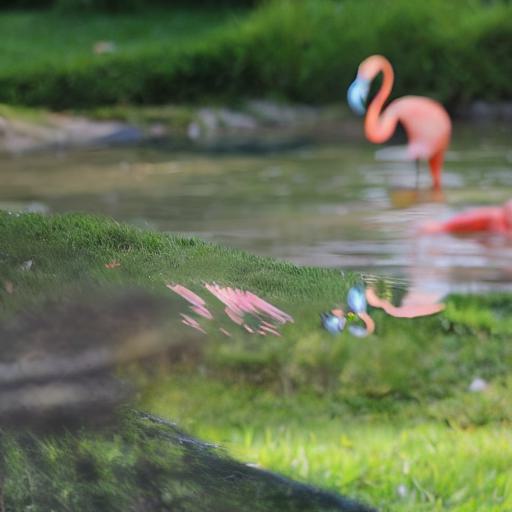}}&
        \frame{\includegraphics[height=\qualheight]{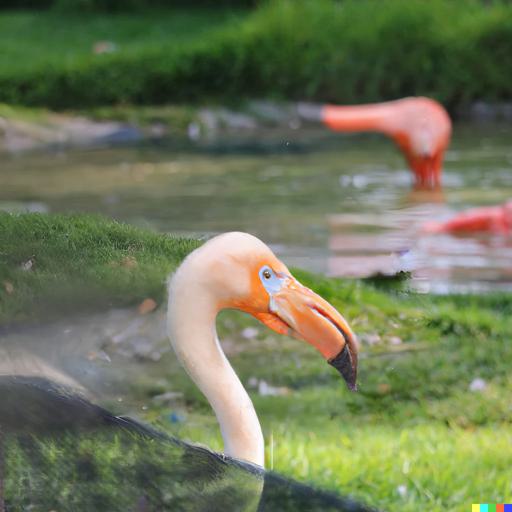}}&
        \frame{\includegraphics[height=\qualheight]{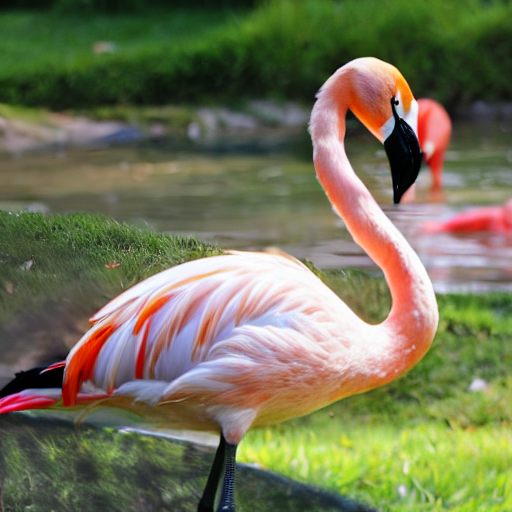}} \\
        
        \raisebox{2.5em}{\rotatebox[origin=c]{90}{\fsh{Mount Fuji}}} &
        \frame{\includegraphics[height=\qualheight]{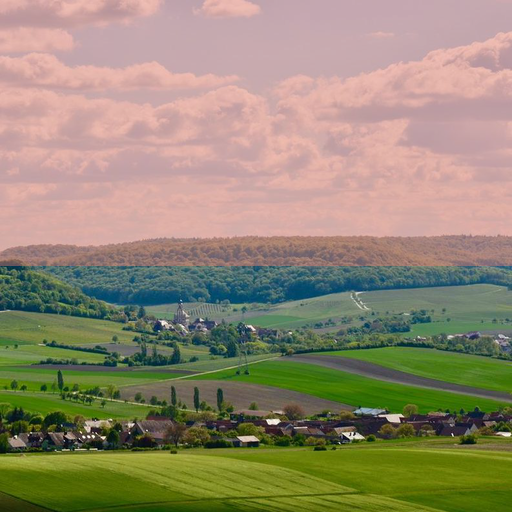}}& \frame{\includegraphics[height=\qualheight]{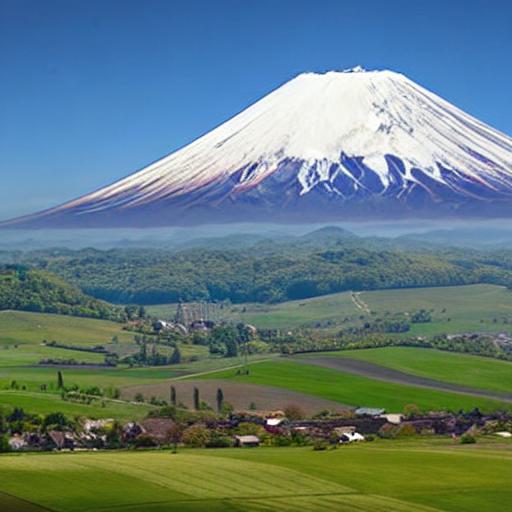}}&
        \frame{\includegraphics[height=\qualheight]{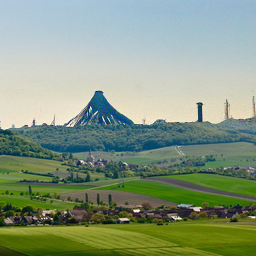}}&
        \frame{\includegraphics[height=\qualheight]{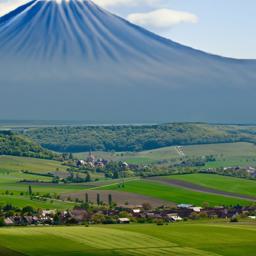}}& \frame{\includegraphics[height=\qualheight]{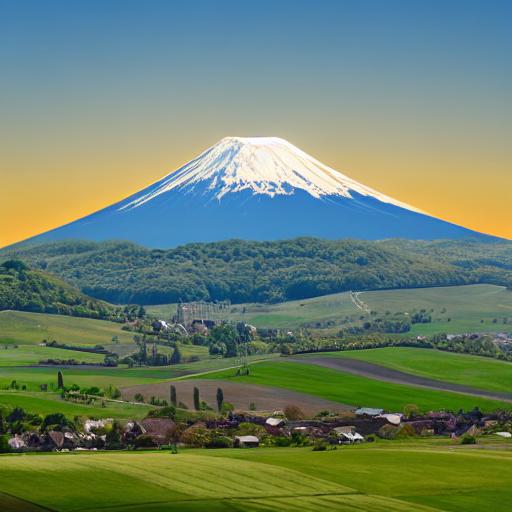}}&
        \frame{\includegraphics[height=\qualheight]{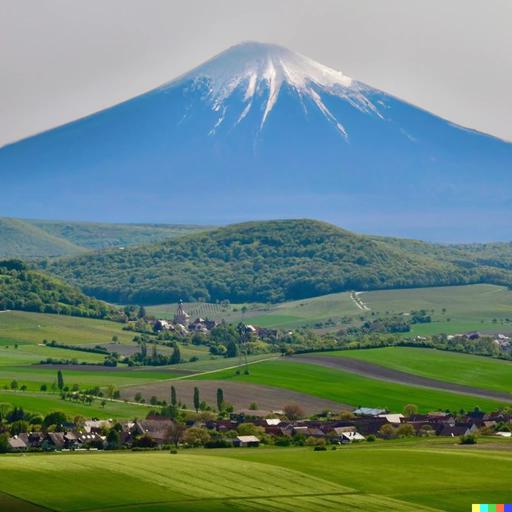}}&
        \frame{\includegraphics[height=\qualheight]{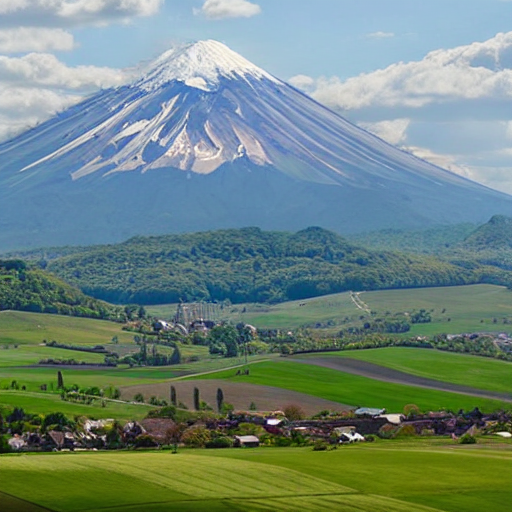}} \\
\end{tabular}
\captionof{figure}{Our method generates high-quality object inpainting results. Different mask precision levels allowing users to either provide exact masks (top row) or to use a rough mask outline (bottom row). Compared to existing methods, our method generates more realistic images, follows accurate masks more closely (top row) and shows better background preservation for coarse masks (bottom row).}
\label{fig:sample_inpainting}
\end{center}
}]
\let\thefootnote\relax\footnotetext{* Work done during internship at Adobe.}

\begin{abstract}
Generic image inpainting aims to complete a corrupted image by borrowing surrounding information, which barely generates novel content. By contrast, multi-modal inpainting provides more flexible and useful controls on the inpainted content, \eg, a text prompt can be used to describe an object with richer attributes, and a mask can be used to constrain the shape of the inpainted object rather than being only considered as a missing area. We propose a new diffusion-based model named SmartBrush for completing a missing region with an object using both text and shape-guidance. While previous work such as DALLE-2 and Stable Diffusion can do text-guided inapinting they do not support shape guidance and tend to modify background texture surrounding the generated object. Our model incorporates both text and shape guidance with precision control. To preserve the background better, we propose a novel training and sampling strategy by augmenting the diffusion U-net with object-mask prediction. Lastly, we introduce a multi-task training strategy by jointly training inpainting with text-to-image generation to leverage more training data. We conduct extensive experiments showing that our model outperforms all baselines in terms of visual quality, mask controllability, and background preservation.
\end{abstract}


\section{Introduction}
Traditional image inpainting aims to fill the missing area in images conditioned on surrounding  pixels, lacking control over the inpainted content. To alleviate this, multi-modal image inpainting offers more control through additional information, \eg class labels, text descriptions, segmentation maps, \etc.
In this paper, we consider the task of multi-modal object inpainting conditioned on both a text description and the shape of the object to be inpainted (see \cref{fig:sample_inpainting}). In particular, we explore diffusion models for this task inspired by their superior performance in modeling complex image distributions and generating high-quality images.

Diffusion models (DMs)~\cite{sohl2015deep,ho2020denoising}, \eg, Stable Diffusion~\cite{rombach2022high}, DALL-E~\cite{ramesh2021zero,ramesh2022hierarchical}, and Imagen~\cite{saharia2022photorealistic} have shown promising results in text-to-image generation. They can also be adapted to the inpainting task by replacing the random noise in the background region with a noisy version of the original image during the diffusion reverse process~\cite{lugmayr2022repaint}. However, this leads to undesirable samples since the model cannot see the global context during sampling~\cite{nichol2021glide}.
To address this, GLIDE~\cite{nichol2021glide} and Stable Inpainting (inpainting specialist v1.5 from Stable Diffusion)~\cite{rombach2022high} randomly erase part of the image and fine-tune the model to recover the missing area conditioned on the corresponding image caption.
However, semantic misalignment between the missing area (local content) and global text description may cause the model to fill in the masked region with background instead of precisely following the text prompt as shown in \cref{fig:sample_inpainting} (``Glide'' and ``Stable Inpainting''). We refer to this phenomenon as \textbf{\emph{text misalignment}}.

An alternative way to perform multi-modal image inpainting is to utilize powerful language-vision models, \eg, CLIP~\cite{radford2021learning}.
Blended diffusion~\cite{avrahami2022blended} uses CLIP to compute the difference between the image embedding and the input text embedding and then injects the difference into the sampling process of a pretrained unconditional diffusion model.
However, CLIP models tend to capture the global and high-level image features, thus there is no incentive to generate objects aligning with the given mask (see ``Blended Diffusion'' in~\cref{fig:sample_inpainting}). We denote this phenomenon as \textbf{\emph{mask misalignment}}.
Another issue for existing inpainting methods is \textbf{\emph{background preservation}} in which case they often produce distorted background surrounding the inpainted object as shown in \cref{fig:sample_inpainting} (bottom row).



To address above challenges, we introduce a precision factor into the input masks, \ie, our model not only takes a mask as input but also information about how closely the inpainted object should follow the mask's shape.
To achieve this we generate different types of masks from fine to coarse by applying Gaussian blur to accurate instance masks and use the masks and their precision type to train the guided diffusion model.
With this setup, we allow users to either use coarse masks which will contain the desired object somewhere within the mask or to provide detailed masks that outline the shape of the object exactly.
Thus, we can supply very accurate masks and the model will fill the entire mask with the object described by the text prompt (see the first row in \cref{fig:sample_inpainting}), while, on the other hand, we can also provide very coarse masks (\eg, a bounding box) and the model is free to insert the desired object within the mask area such that the object is roughly bounded by the mask.

One important characteristic, especially for coarse masks such as bounding boxes, is that we want to keep the background within the inpainted area consistent with the original image.
To achieve this, we not only encourage the model to inpaint the masked region but also use a regularization loss to encourage the model to predict an instance mask of the object it is generating.

At test time we replace the coarse mask with the predicted mask during sampling to preserve background as much as possible which leads to more consistent results (second row in \cref{fig:sample_inpainting}).

We evaluate our model on several challenging object inpainting tasks and show that it achieves state-of-the-art results on object inpainting across several datasets and examples.
Our model offers more flexibility due to the mask precision control, which offers users to specify how closely they want the model to follow a given mask.
Due to our foreground mask prediction during sampling, our model is much better at preserving background within the inpainted areas than other baselines, leading to more realistic results, especially for coarser masks such as bounding boxes.
Our user study shows that users prefer the outputs of our model as compared to DALLE-2 and Stable Inpainting across several axes of evaluation such as shape, text alignment, and realism.
To summarize our contributions:
\begin{itemize}
    \item We introduce a text and shape guided object inpainting diffusion model, which is conditioned on object masks of different precision, achieving a new level of control for object inpainting.
    \item To preserve the image background with coarse input masks, the model is trained to predict a foreground object mask during inpainting for preserving original background surrounding the synthesized object.
    \item Instead of training with random masks and text captions that describe the entire images, we use instance segmentation masks and train our model with local text descriptions of the inpainted area. 
    \item We propose a multi-task training strategy by jointly training object inpainting with text-to-image generation to leverage more training data. 
\end{itemize}

\section{Related Work}

\textbf{Diffusion Models}
Diffusion models (DMs)~\cite{sohl2015deep,ho2020denoising} learn the data distribution by inverting a Markov noising process, and they have gained wide attention recently due to their stability and superior performance in image synthesis as compared to GANs. Given a clean image $x_0$, the diffusion process adds noise to the image at each step $t$, obtaining a set of noisy latent $x_t$. Then, the model is trained to recover the clean image $x_0$ from $x_t$ in the backward process.  DMs have shown appealing results in different tasks, \eg, unconditional image generation~\cite{ho2022cascaded,ho2020denoising, song2019generative,song2020denoising}, text-to-image generation~\cite{ramesh2021zero,ramesh2022hierarchical,saharia2022photorealistic, rombach2022high}, video generation~\cite{ho2022imagen}, image inpainting \cite{nichol2021glide, avrahami2022blended,avrahami2022blended2,lugmayr2022repaint}, image translation~\cite{wang2022pretraining,meng2021sdedit,zhao2022egsde}, and image editing~\cite{hertz2022prompt,couairon2022diffedit,kawar2022imagic}.


\textbf{Text-Guided Image Inpainting}
Taking advantage of the recent success of diffusion-based text-to-image generation models,  an intuitive adaptation from a text-to-image generation to text-guided inpainting is to replace the pure random noise with the noisy background outside the mask region. However, this leads to strong artifacts, \eg, generating partial objects or inconsistent content in the background.
To address this problem, GLIDE~\cite{nichol2021glide} further finetune a pre-trained text-to-image model toward the inpainting task. It first generates a random mask and then provides the masked image and mask as additions to the diffusion model, which learns to utilize the information outside of the mask region. Blended  diffusion~\cite{avrahami2022blended} adapts from a pre-trained unconditional diffusion model and encourages the output to align with the text prompt using the CLIP score.  
Repaint~\cite{lugmayr2022repaint} builds on a pre-trained unconditional diffusion model and proposes to resample in each reverse step, but it doesn't support text input. Some recent works also endeavored to tackle image editing tasks, \eg, Prompt2Prompt~\cite{hertz2022prompt} allows partial modification on the original prompt such that the newly generated image will be partially edited correspondingly, while it is difficult to control object shape and target regions, especially if the image content becomes complicated. DiffEdit~\cite{couairon2022diffedit} follows the spirit of Prompt2Prompt but derives masks from the difference before and after modifying the prompt. PaintbyWord~\cite{bau2021paint} pairs the large-scale GAN with a full-text image retrieval network to enable multi-modal image editing. However, due to the structure of GAN, it cannot specifically modify the region given by the mask. TDANet \cite{zhang2020text} proposes a dual attention mechanism to exploit the text features about the masked region by comparing text with the corrupted image and its counterpart.


\section{Preliminary: Diffusion Model}
Given an input image $x_0$, we apply a forward diffusion Markov process to add noise to the image over a number of time steps $t$ with scheduled variance $\beta_t$: 
\begin{align}
    q(x_t|x_{t-1}) &= \mathcal{N}\left(\sqrt{1-\beta_t} x_t, \beta_t \text{I}\right)\\
    q(x_{1:T}|x_0) &= \prod q(x_t|x_{t-1}), \nonumber
    \label{eq:forward}
\end{align}
where $T$ is the total number of steps. If $T\rightarrow\infty$, the output $x_T$ will be isotropic Gaussian. The defined Markov process allows us to get $x_t$ in a closed form
\begin{align}
    x_t &= \sqrt{{\alpha}_t}x_{t-1} + \sqrt{1-{\alpha}_t}\epsilon_{t-1}\\ \nonumber
   &= \sqrt{\bar{\alpha}_t}x_0 + \sqrt{1-\bar{\alpha}_t}\epsilon,
\end{align}
where $\alpha_t=1-\beta_t$, $\bar{\alpha}_t=\prod_{i=1}^t \alpha_i$, $\epsilon_t \sim \mathcal{N}(0, \text{I})$.

To generate images from random noise, we need to invert above diffusion process, \ie, learning $q(x_{t-1}|x_t)$ that is also a Gaussian when $\beta_t$ is small enough. However, $q(x_{t-1}|x_t)$ is unknown since it is inaccessible to the true distribution of $x_0$. Thus, we train a neural network $p_{\theta}$ to approximate the conditional distribution. 
\begin{align}
    p_{\theta}(x_{t-1}|x_t) &= \mathcal{N}\left(\mu_{\theta}(x_t, t), \Sigma_{\theta}(x_t, t)\right),
\end{align}
where $\mu_{\theta}$ is trained to predict $x_{t-1}=\frac{1}{\sqrt{\alpha_t}}\left(x_t - \frac{1-\alpha_t}{\sqrt{1-\bar{\alpha}_t}\epsilon_t}\right)$, which is derived from \cref{eq:forward}. Since we already have $x_t$ during training, we can train a network $\epsilon_{\theta}$ to predict $\epsilon_t$ instead of training $\mu_{\theta}$ \cite{ho2020denoising}. We obtain the objective for training the diffusion model. 
\begin{align}
    \mathcal{L} &= \mathbb{E}_{t\sim [1, T], x_0, \epsilon_t}\left\|\epsilon_t - \epsilon_{\theta}(x_t, t) \right\|_2^2
\end{align}
At test time, we start from a random noise $x_T\sim \mathcal{N}(0, \text{I})$ and then iteratively apply the model $\epsilon_{\theta}$ to obtain $x_{t-1}$ from $x_t$ until $t=0$. We may employ more efficient sampling techniques like DDIM~\cite{song2020denoising} and PNDM~\cite{liu2022pseudo} to speed up the sampling, and adopt classifier free guidance~\cite{ho2022classifier} to improve the sample quality. 

As for conditional diffusion models, \eg, text-to-image and inpainting models, conditional information can be fed into the network $\epsilon_{\theta}$ without changing the loss function. The model will learn to utilize the conditions to generate high quality conditional images.

\begin{figure*}
    \centering
    \includegraphics[scale=0.43]{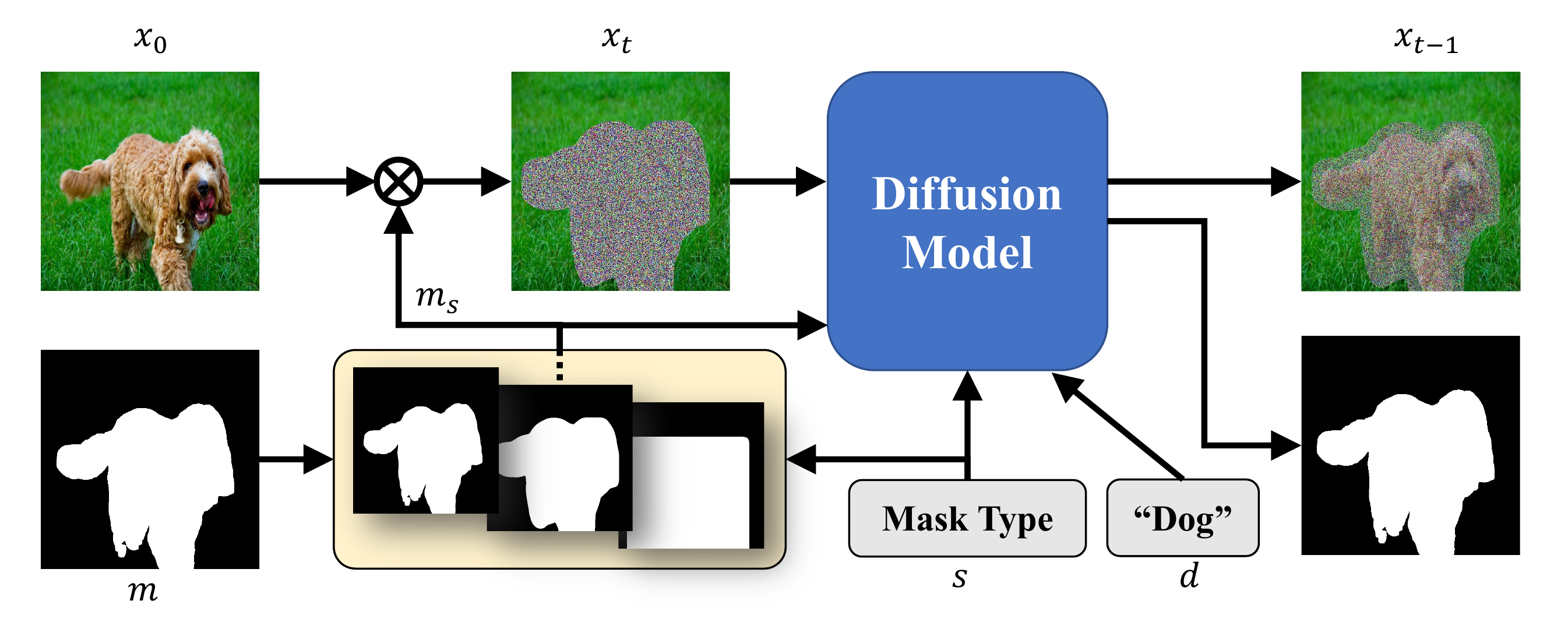}
    \caption{Text and shape guided object inpainting. Given an image $x_0$, accurate mask $m$ and object description $d$, we transform the mask $m$ to different precision levels (from accurate to coarse) as $m_s$. We add noise in the masked region to provide rich background information to the diffusion model and train the model to predict the added noise as well as the accurate mask $m$. During inference, we apply the diffusion model repeatedly until $t=0$.}
    \label{fig:my_label}
\end{figure*}

\section{Our Approach}
Given an image $x$, text prompt $d$ and a binary mask $m$ to indicate which region of $x$ we should modify, our goal is to generate an image $\tilde{x}$ such that the background of $\tilde{x}$ is the same as input $x$ while the generation in the masked region $\tilde{x}\odot m$ aligns well with the text prompt $d$ and the mask $m$.

\subsection{Text and Shape Guided Diffusion}
Existing inpainting models randomly erase part of the images and are trained to inpaint the erased region.
As a result, the randomly erased region may contain only parts of an object or contain areas of background around a given object. 
Therefore, we propose to utilize the text and shape information from existing instance or panoptic segmentation datasets.
These datasets contain annotated masks $\{m_i\}_{i=1}^N$ where $N$ is the number of annotations and each masked region $x\odot m_i$ contains only one object.
For each mask we also have a corresponding class label $c_i$, \eg, \emph{hat} or \emph{cat}. 

In the forward process, we randomly draw a segmentation mask $m$ and its corresponding class text label $c$ for image $x$.
We define $x_0=x$ and only add noise in the masked region instead of all pixels:
\begin{align}
    &\tilde{x}_t = \sqrt{\bar{\alpha}_t} x_0 + \sqrt{1-\bar{\alpha}_t}\epsilon\\ \nonumber
    &x_t = \tilde{x}_t \odot m + x_0 \odot (1-m),
\end{align}
where $\epsilon\sim \mathcal{N}(\textbf{0}, \textbf{I})$ and $t$ is the timestep in the forward process.
We use $x_t$, $m$, and $c$ as input to the model so it can learn to utilize the clean background information and learn to recover the masked region $x_0\odot m$.
This ensures that generated objects in the foreground $m$ are consistent with the background. 
%
Following \cite{ho2020denoising} we train a network $\epsilon_{\theta}$ to predict the noise $\epsilon$ from the noisy $x_t$:
\begin{align}
    \mathcal{L}_{\text{DM}} = \mathbb{E}_{\epsilon \sim \mathcal{N}(0, I)}\left[\|\epsilon - \epsilon_\theta(x_t, t, m, c)\|_2^2\right].
\end{align}
In the inference phase, we generate random Gaussian noise in the masked region $x_T=\epsilon \odot m + x_0 \odot (1-m)$, where $T$ is the number of sampling steps. Then we reverse the diffusion process and obtain the inpainted result $x_0$.

\subsection{Shape Precision Control}
Our training masks come from the segmentaion annotations and thus are accurate instance masks.
Training the model with these masks will encourage the model to exactly follow the shape of the input mask at test time.
To allow users to provide masks that are either accurate (e.g., in the shape of a cat) or coarse (e.g., a bounding box) we propose to generate masks with different precision.
To achieve this, we randomly augment the masks during training to degrade the shape of the original mask.
Specifically, given an accurate instance mask $m$, we use a mask precision indicator $s\sim [0, S]$ and define a set of parameters for each indicator:
\begin{align}
    m_s = \text{GaussianBlur}(m, k_s,\sigma_s),
\end{align}
where $k_s$ denotes Gaussian kernel size, and $\sigma_s$ is standard deviation of the kernel.
If $s=0$, the mask stays unchanged and corresponds to the accurate instance mask from the dataset annotation.
When $s=S$, the mask $m_s$ is a bounding box of the instance mask $m$, and it loses all detailed shape information.
During training, for each training sample (object), we employ a set of masks $\{m_s, s\}$ from fine to coarse and condition the diffusion model on the precision indicator $s$:
\begin{align}
     \mathcal{L}_{\text{seg-DM}} = \mathbb{E}_{\epsilon \sim \mathcal{N}(0, I)}\left[\|\epsilon - \epsilon_\theta(x_t, t, m_s, c, s)\|_2^2\right].
     \label{eq:l_seg}
\end{align}
Through this, we can control whether the generated object should align with the input mask by specifying different mask precision indicators $s$.
We present a sample of masks in Fig. \ref{fig:mask_precision}.

\subsection{Background Preservation}
During inference, the diffusion model will denoise the masked region and generate objects according to the given text prompt.
As a result, the background in the masked region will be changed if the input masks are coarse.
For example, the model may generate a cat in the given square box mask region but the other pixels in the square box region will also be changed.
Ideally we would like to preserve the background, however, this is challenging since we do not know where in the coarse mask the model will generate the desired object.

We address this challenge by utilizing the information of mask precision.
Specifically, we train our diffusion network to also predict an accurate instance mask $m$ from the coarse input version $m_s$:
\begin{align}
    \mathcal{L}_{\text{prediction}}= H(\epsilon_\theta(m_s), m),
    \label{eq:l_pred}
\end{align}
where $H$ can be any suitable criterion for segmentation.
We choose to use the DICE loss, \ie, $H(X,Y)=1-\frac{2|X\cap Y|}{|X|+|Y|}$.
For this, we simply add an extra output channel to our diffusion model which contains the instance mask prediction.

During inference, we are able to predict where the object is generated inside the coarse mask $m_s$ using the diffusion model's prediction.
We first feed a coarse mask $m_s$ into the diffusion model and switch to using the predicted mask to perform denoising.
With the predicted mask, we know where the object is generated within the masked region which helps to preserve background information around the generated object. 


\subsection{Training Strategy}
\label{subsec:training}
Combining \cref{eq:l_seg,eq:l_pred}, our final training objective can be expressed as follows.
\begin{align}
    \mathcal{L}_{\text{total}} = \mathcal{L}_{\text{seg-DM}} + \lambda \mathcal{L}_{\text{prediction}},
    \label{eq:total_loss}
\end{align}
where $\lambda$ is a hyper-parameter which balances the two losses. In our experiment, $\lambda=0.01$.

Our model can be built based on pre-trained text-to-image generation models, \eg, Stable Diffusion and Imagen, to speed up the training process. In the experiments, we finetune based on the Stable Diffusion text-to-image model v1.2~\footnote{https://github.com/CompVis/stable-diffusion} with our conditions (\cref{fig:my_label}) and loss function $\mathcal{L}_{\text{total}}$ (\cref{eq:total_loss}). 
To align text descriptions with the local mask content, avoiding text misalignment as aforementioned, we train with the training split of OpenImages v6~\footnote{https://storage.googleapis.com/openimages/web/index.html}, which has segmentation and corresponding labels that can serve as local descriptions. From our empirical study, such categorical text would degrade the generation quality from long sentences. Therefore, we employ the BLIP model~\cite{li2022blip} to collect richer and longer captions for those local segments. During the training, we randomly pair the segmentation label or BLIP caption to the corresponding mask. Therefore, the model can handle both single word text and short phrase well during the inference. 

\textbf{Multi-task Training} In addition, to leverage more training data and handle more diverse text descriptions and image contents, beyond the domain of the segmentation dataset, we propose a multi-task training strategy by jointly training our main task and the foundational text-to-image generation task, using image/text paired data from LAION-Aesthetics v2 5+ subset~\cite{schuhmann2021laion} following Stable Diffusion~\cite{rombach2022high}. For text-to-image, we set the input mask to cover the entire image, and treat it as a special inpainting case. As demonstrated in \cref{sec:experiments}, our final model trained with all these components significantly outperforms state-of-the-art methods in terms of visual quality of generated objects, as well as their consistency to text description and mask shape.

\section{Experimental Evaluation}
\label{sec:experiments}
\subsection{Experimental Setup}
We set $\lambda=0.01$ in the total loss function~\cref{eq:total_loss} and batch size to be 1024. Following the training strategy discussed in \cref{subsec:training}, we train the inpainting task and text-to-image generation task with the probability of 80\% and 20\%, respectively. Our model was trained around 20K steps on 8 A100 GPUs. As a reference, Stable Inpainting takes 256 A100 GPUs around 440K steps.

\setlength{\tabcolsep}{1.5pt}
\begin{figure*}
    \centering
    \def\fsh{\footnotesize}
    \def\fsc{\small}
    \setlength{\tabcolsep}{1pt}
    \def\qualwidth{2.3cm}
    \begin{tabular}{cccccccc}
    &\fsc{Input+Mask} & \fsc{Blended Diffusion}  & \fsc{Stable Diffusion} & \fsc{GLIDE} & \fsc{Stable Inpainting} & \fsc{DALLE-2} & \fsc{SmartBrush} \\
        \raisebox{3em}{\rotatebox[origin=c]{90}{\fsh{teddy bear}}}&\includegraphics[width=\qualwidth]{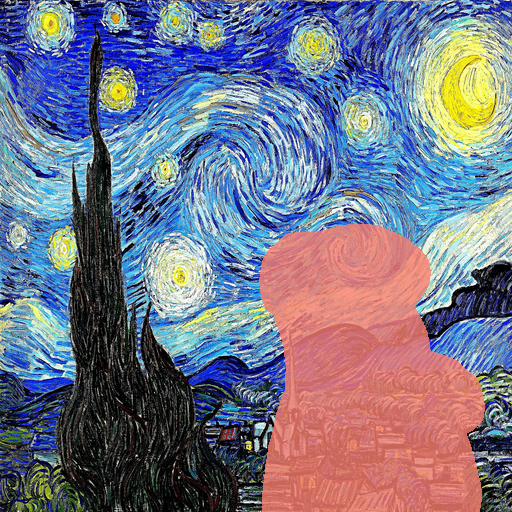} &
         \includegraphics[width=\qualwidth]{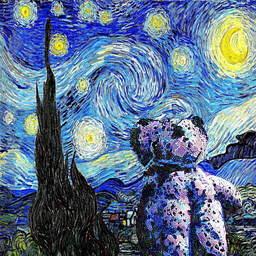}
          &
          \includegraphics[width=\qualwidth]{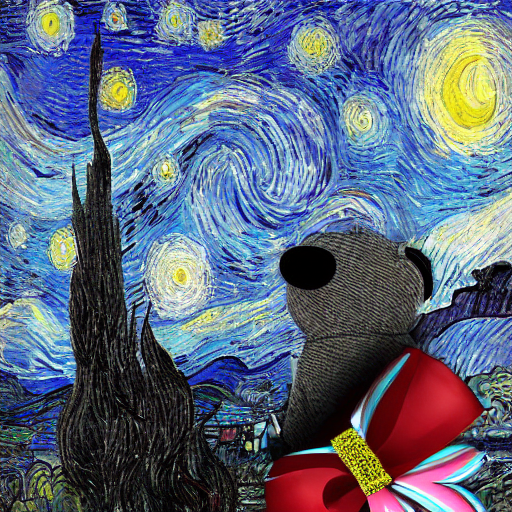}
         &
          \includegraphics[width=\qualwidth]{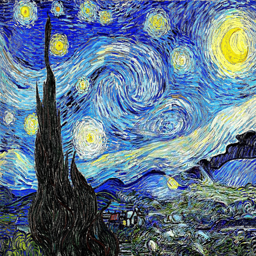}
          &
           \includegraphics[width=\qualwidth]{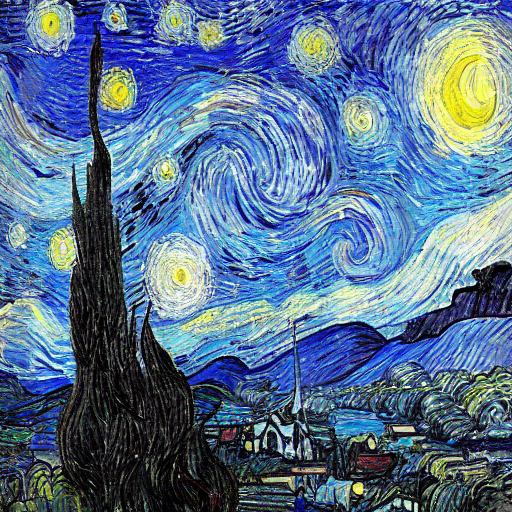}&
            \includegraphics[width=\qualwidth]{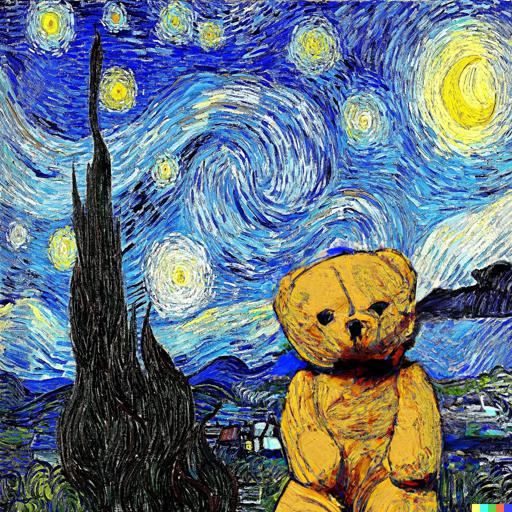}&
             \includegraphics[width=\qualwidth]{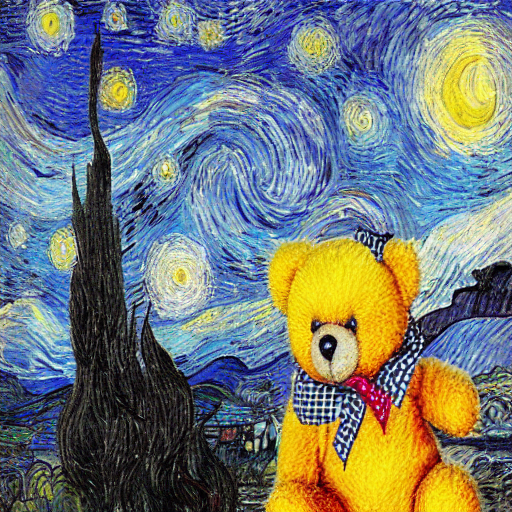}\\
              \raisebox{3em}{\rotatebox[origin=c]{90}{\fsh{persian cat}}}&\includegraphics[width=\qualwidth]{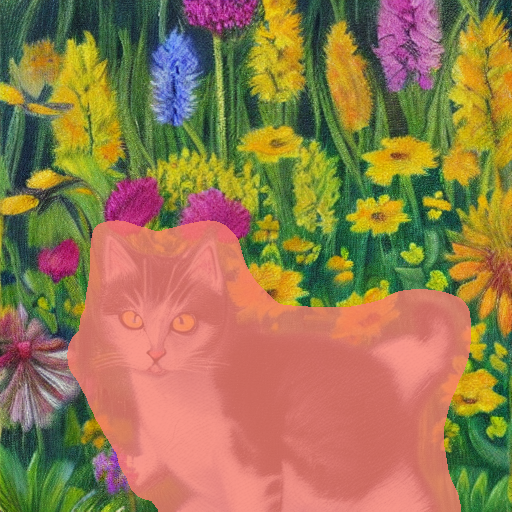} &
         \includegraphics[width=\qualwidth]{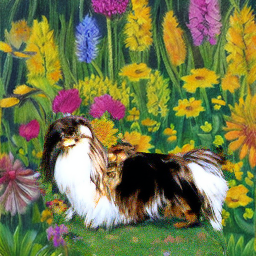}
          &
          \includegraphics[width=\qualwidth]{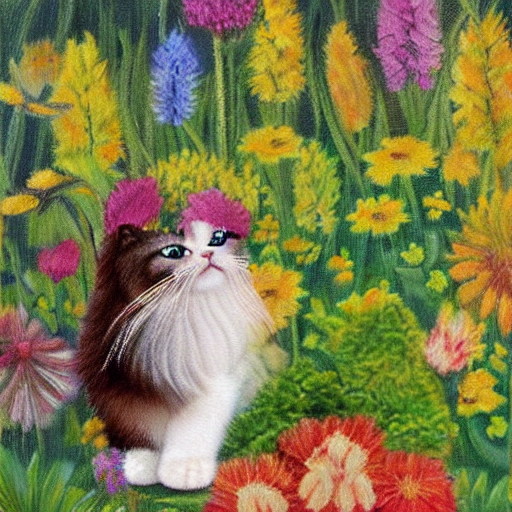}
         &
          \includegraphics[width=\qualwidth]{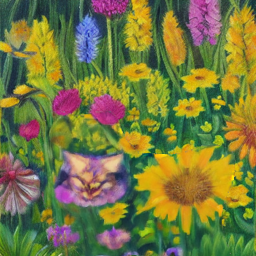}
          &
           \includegraphics[width=\qualwidth]{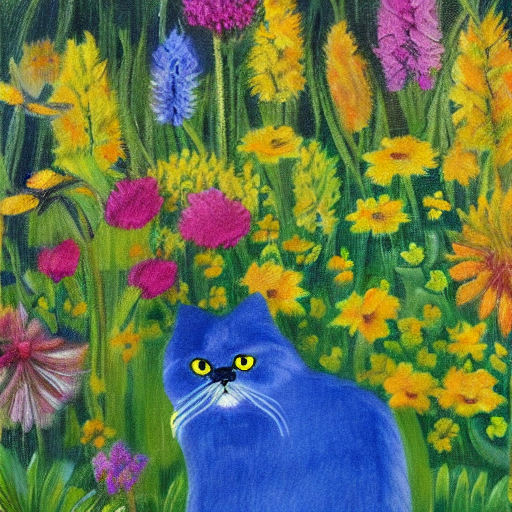}&
            \includegraphics[width=\qualwidth]{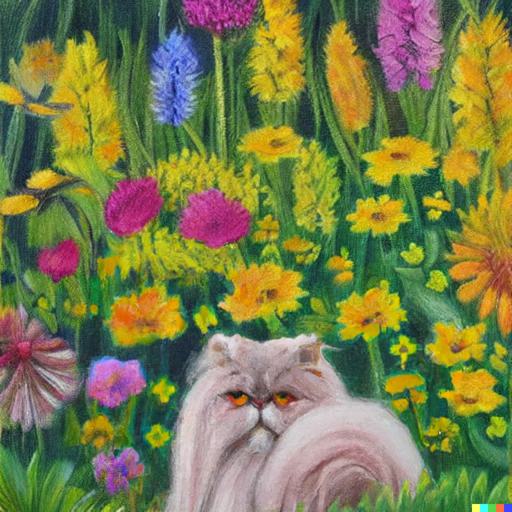}&
             \includegraphics[width=\qualwidth]{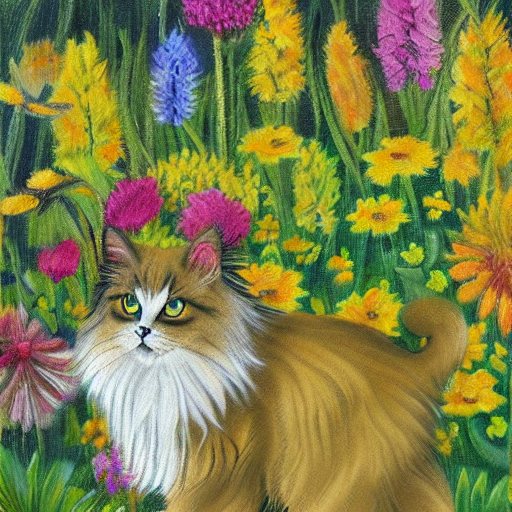}\\
                \raisebox{3em}{\rotatebox[origin=c]{90}{\fsh{table lamp}}}&\includegraphics[width=\qualwidth]{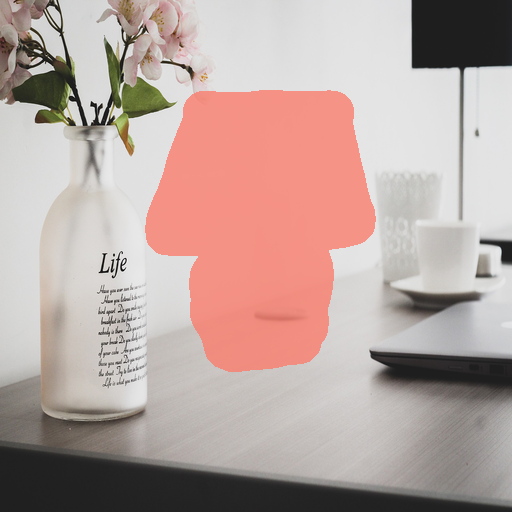} &
         \includegraphics[width=\qualwidth]{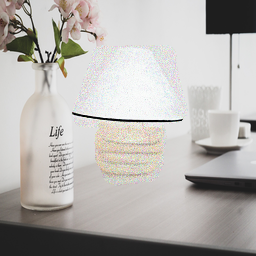}
          &
          \includegraphics[width=\qualwidth]{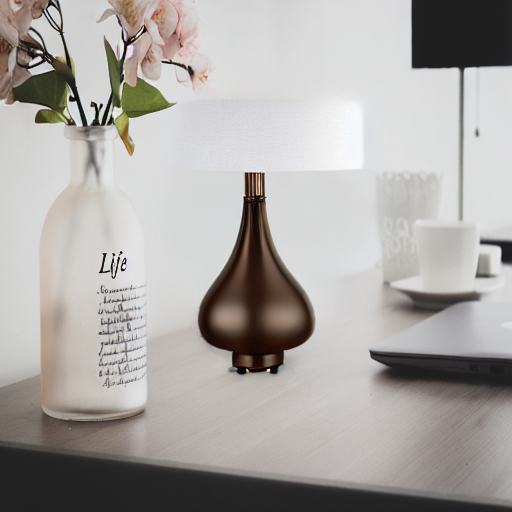}
         &
          \includegraphics[width=\qualwidth]{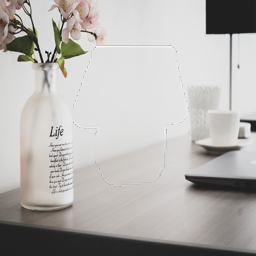}
          &
           \includegraphics[width=\qualwidth]{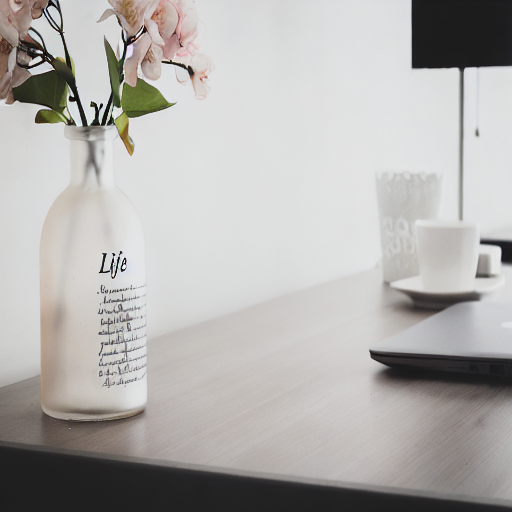}&
            \includegraphics[width=\qualwidth]{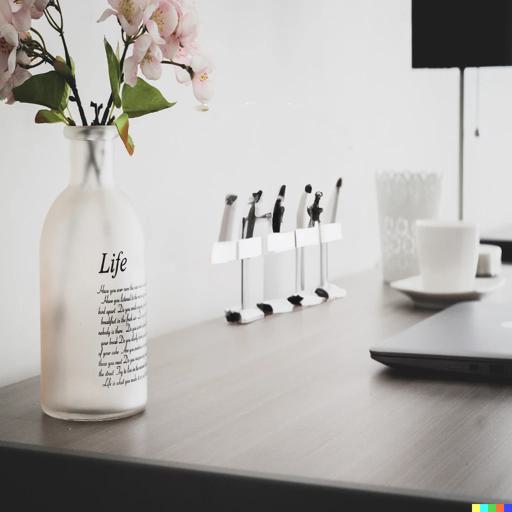}&
             \includegraphics[width=\qualwidth]{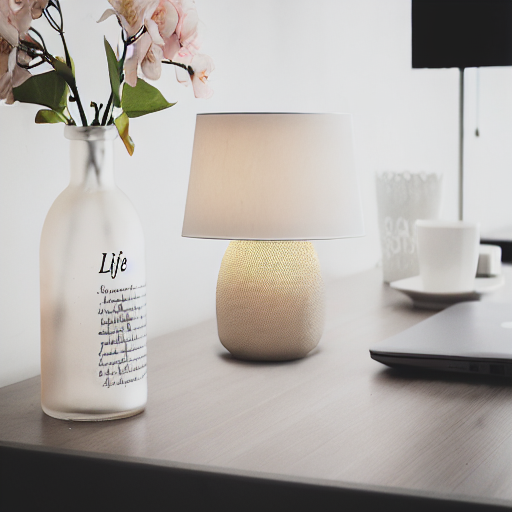}\\
              \raisebox{3em}{\rotatebox[origin=c]{90}{\fsh{corgi}}}&\includegraphics[width=\qualwidth]{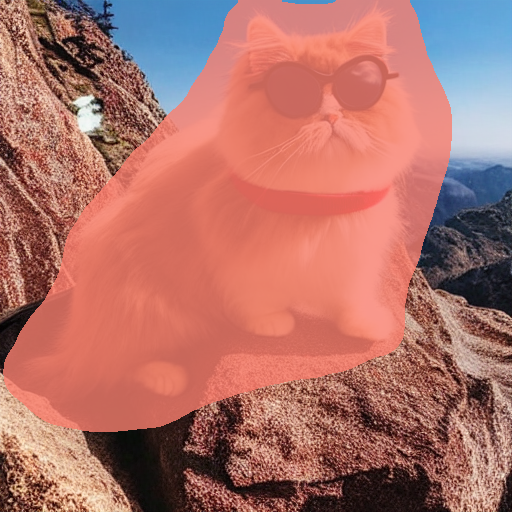} &
         \includegraphics[width=\qualwidth]{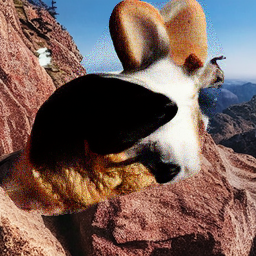}
          &
          \includegraphics[width=\qualwidth]{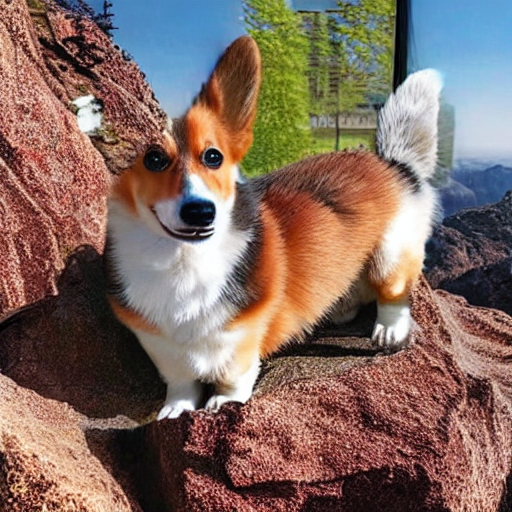}
         &
          \includegraphics[width=\qualwidth]{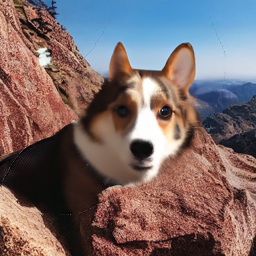}
          &
           \includegraphics[width=\qualwidth]{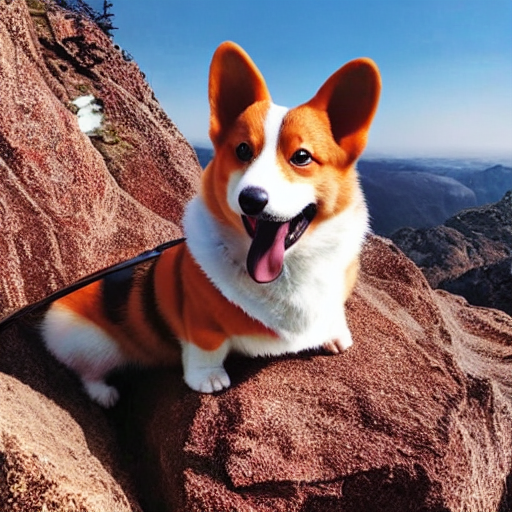}&
            \includegraphics[width=\qualwidth]{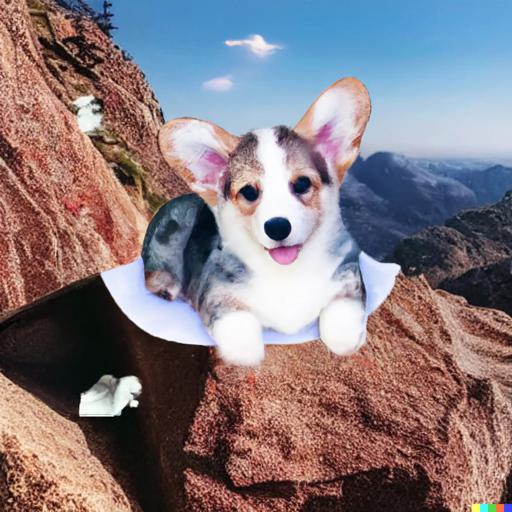}&
             \includegraphics[width=\qualwidth]{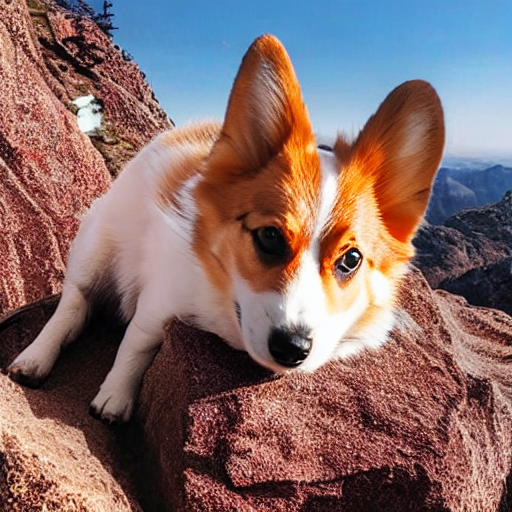}\\
              \raisebox{3em}{\rotatebox[origin=c]{90}{\fsh{\begin{tabular}[x]{@{}c@{}}fuzzy panda wearing\\cowboy hat playing guitar\end{tabular}}}}&\includegraphics[width=\qualwidth]{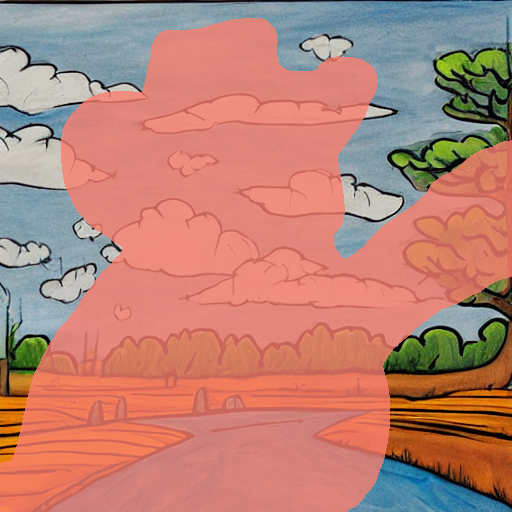} &
         \includegraphics[width=\qualwidth]{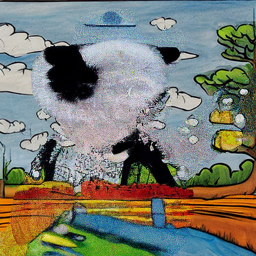}
          &
          \includegraphics[width=\qualwidth]{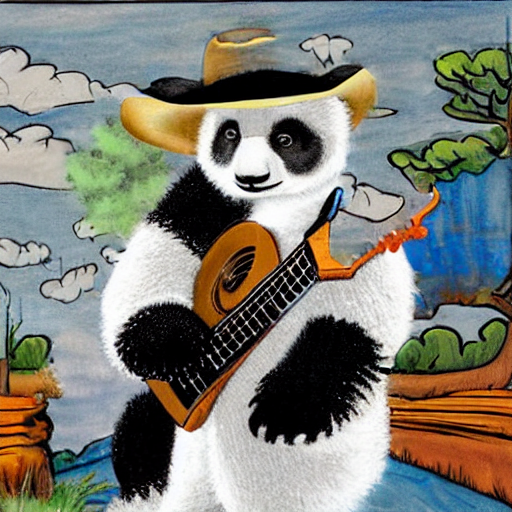}
         &
          \includegraphics[width=\qualwidth]{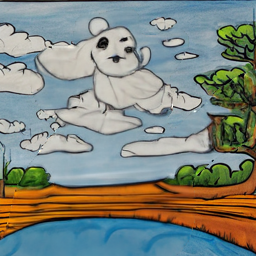}
          &
           \includegraphics[width=\qualwidth]{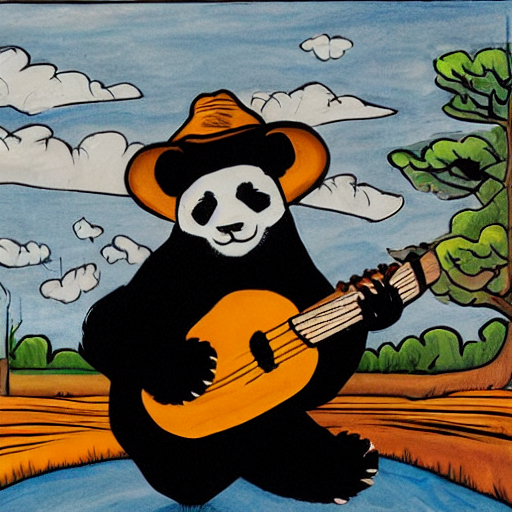}&
            \includegraphics[width=\qualwidth]{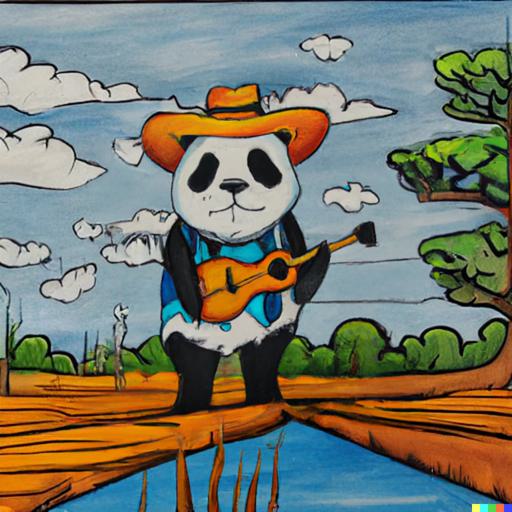}&
             \includegraphics[width=\qualwidth]{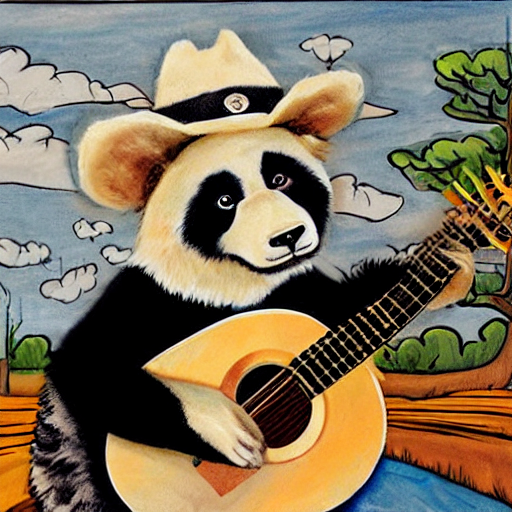}\\
              \raisebox{3em}{\rotatebox[origin=c]{90}{\fsh{\begin{tabular}[x]{@{}c@{}}blue and white striped\\turtleneck sweater\end{tabular}}}}&\includegraphics[width=\qualwidth]{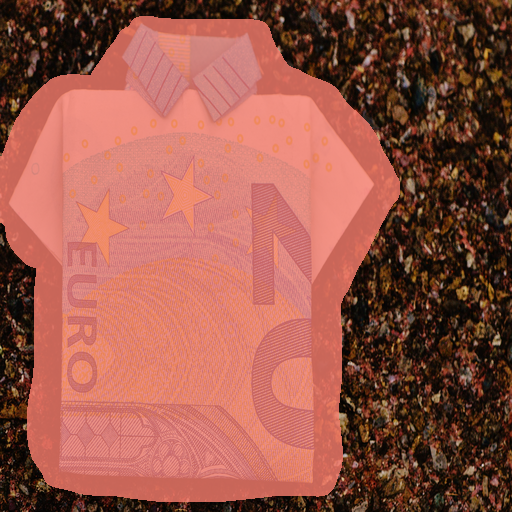} &
         \includegraphics[width=\qualwidth]{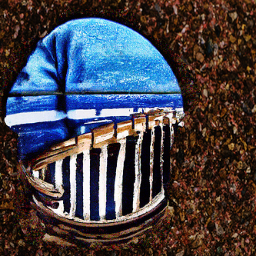}
          &
          \includegraphics[width=\qualwidth]{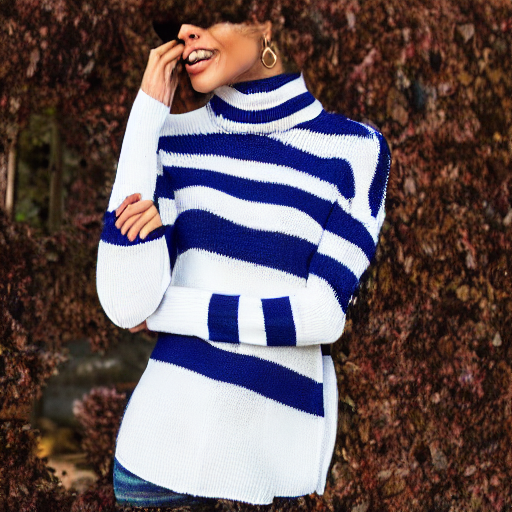}
         &
          \includegraphics[width=\qualwidth]{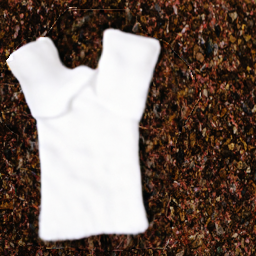}
          &
           \includegraphics[width=\qualwidth]{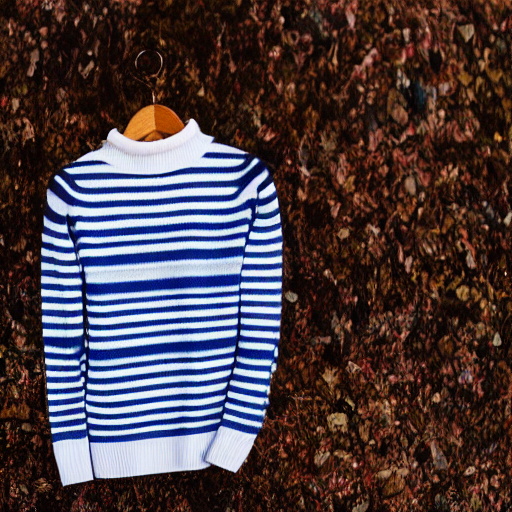}&
            \includegraphics[width=\qualwidth]{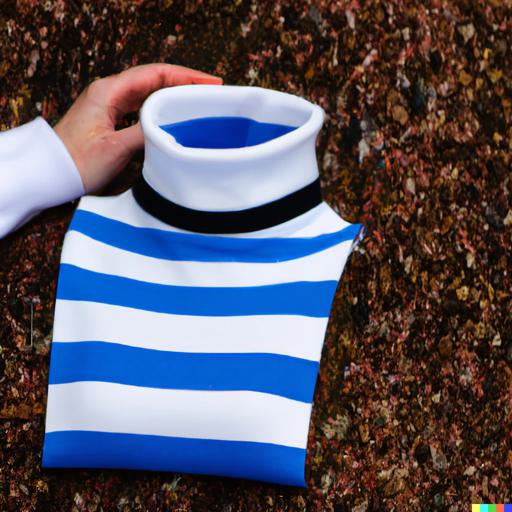}&
             \includegraphics[width=\qualwidth]{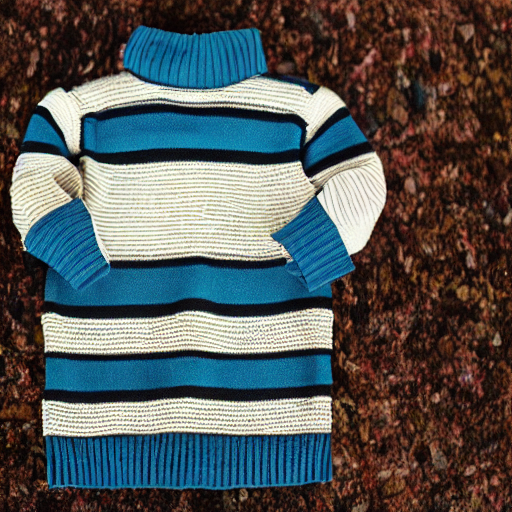}\\ \\
             \hdashline \\
              \raisebox{3em}{\rotatebox[origin=c]{90}{\fsh{buildings}}}&\includegraphics[width=\qualwidth]{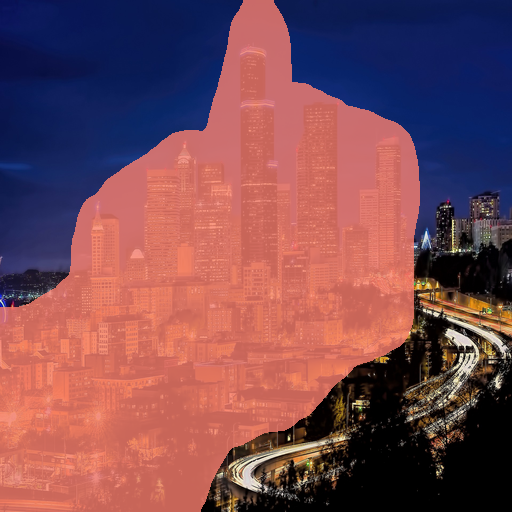} &
         \includegraphics[width=\qualwidth]{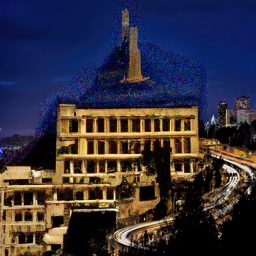}
          &
          \includegraphics[width=\qualwidth]{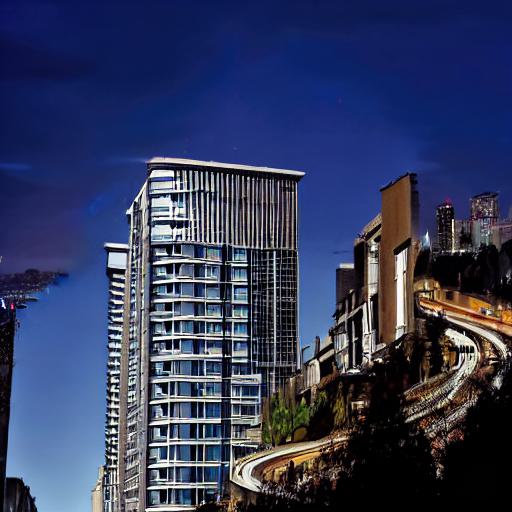}
         &
          \includegraphics[width=\qualwidth]{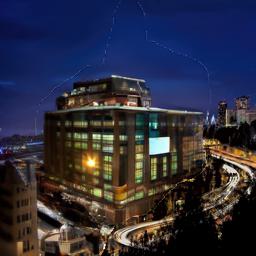}
          &
           \includegraphics[width=\qualwidth]{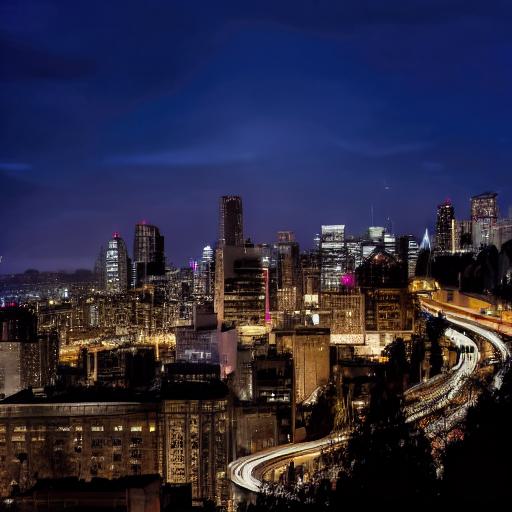}&
            \includegraphics[width=\qualwidth]{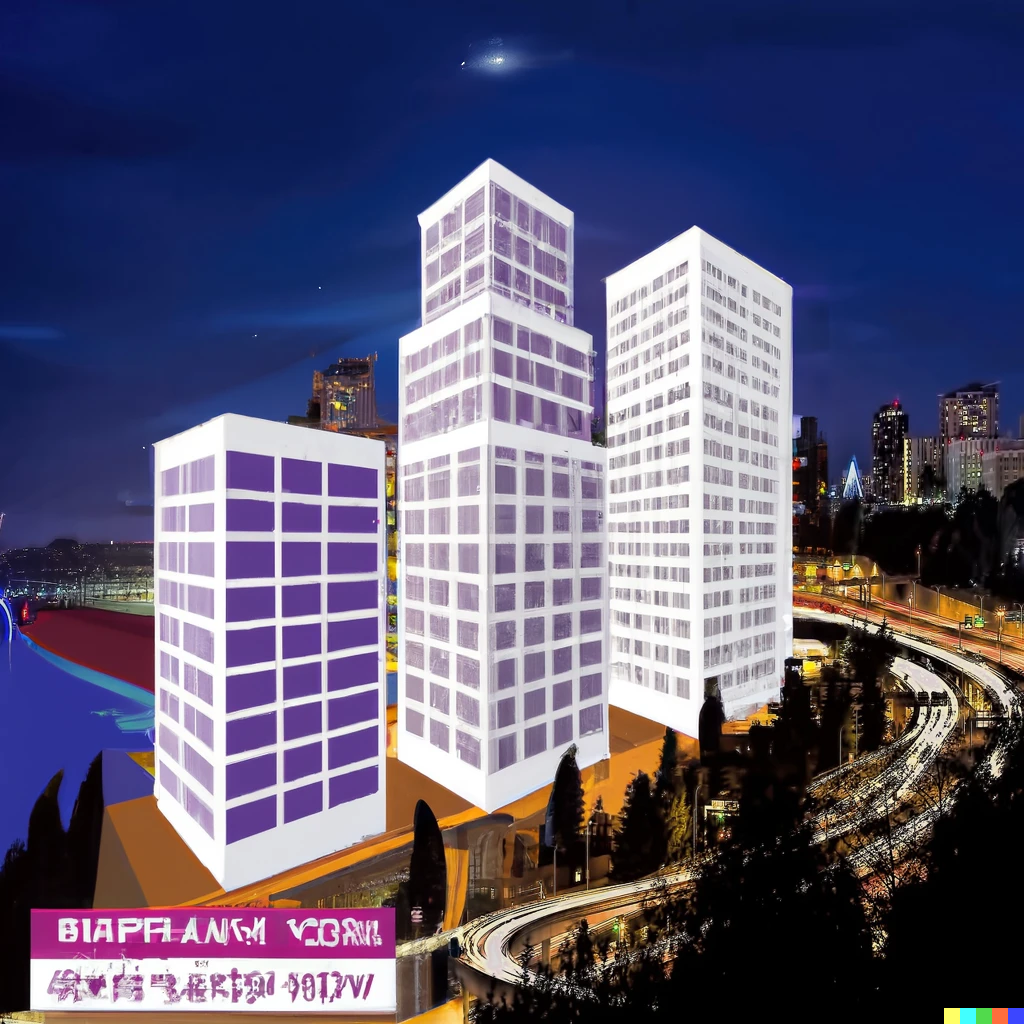}&
             \includegraphics[width=\qualwidth]{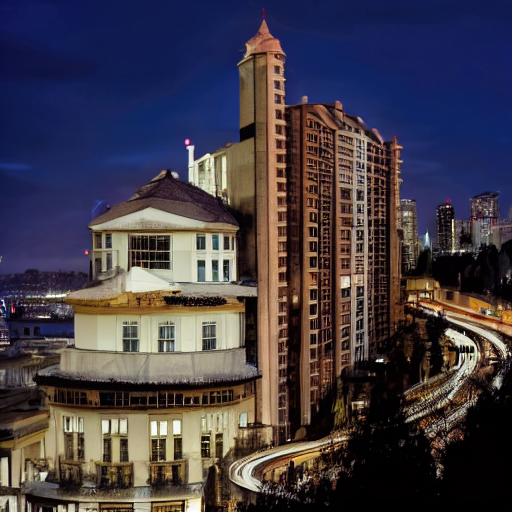}\\
              \raisebox{3em}{\rotatebox[origin=c]{90}{\fsh{\begin{tabular}[x]{@{}c@{}}sunset\\mountain meadow\end{tabular}}}}&\includegraphics[width=\qualwidth]{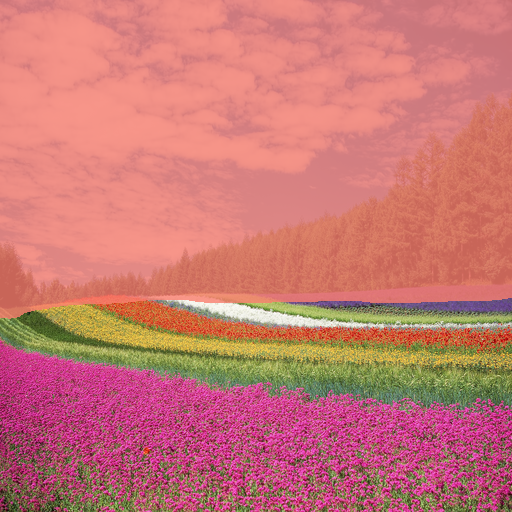} &
         \includegraphics[width=\qualwidth]{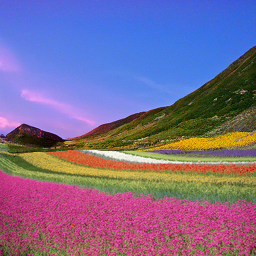}
          &
          \includegraphics[width=\qualwidth]{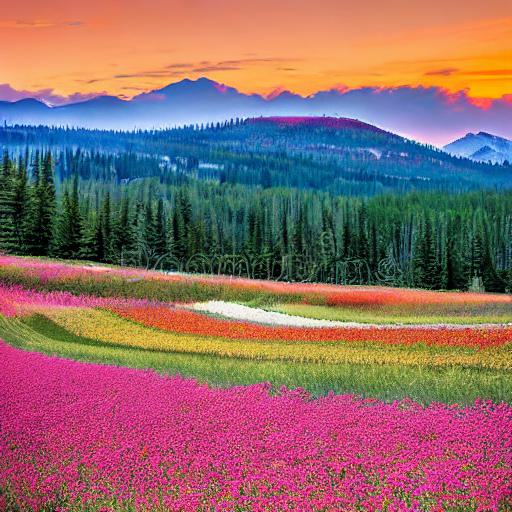}
         &
          \includegraphics[width=\qualwidth]{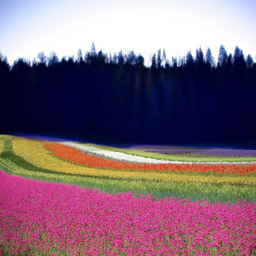}
          &
           \includegraphics[width=\qualwidth]{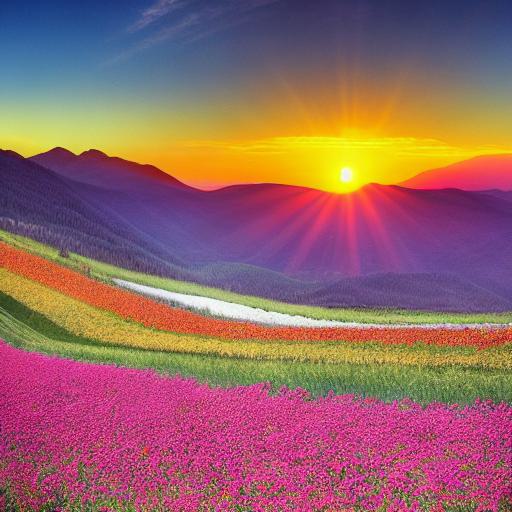}&
            \includegraphics[width=\qualwidth]{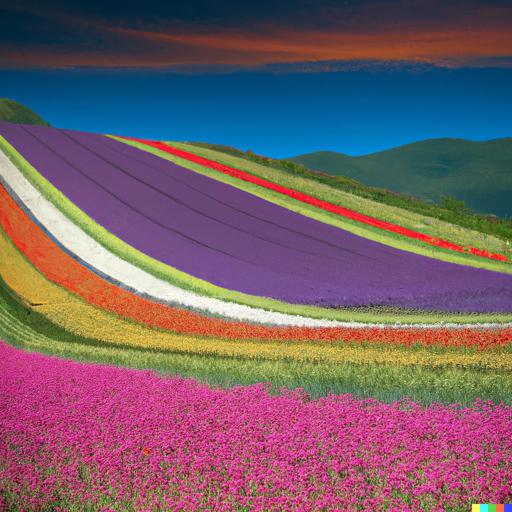}&
             \includegraphics[width=\qualwidth]{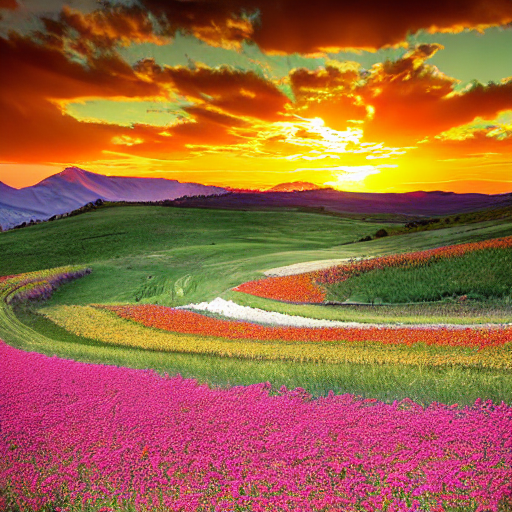}\\
    \end{tabular}
    \caption{Comparison of text and shape guided inpainting.}
    \label{fig:object_generation}
\end{figure*}

\setlength{\tabcolsep}{5pt}
\begin{table*}[]
    \centering
     \caption{Text-guided object inpainting with bounding box mask.}
    
    \begin{tabular}{lcccp{3em}ccc}
    \toprule
    & \multicolumn{3}{c}{OpenImages} && \multicolumn{3}{c}{MSCOCO}\\
    &Local FID $\downarrow$ & CLIP Score $\uparrow$ &FID $\downarrow$ && Local FID $\downarrow$ & CLIP Score $\uparrow$ & FID $\downarrow$\\ \midrule
         Blended Diffusion \cite{avrahami2022blended}& 29.16&0.265&11.05&&41.43&0.251&12.68 \\
         GLIDE \cite{nichol2021glide} &22.45&0.252&9.70&&30.72&0.241&9.32\\ 
         Stable Diffusion \cite{rombach2022high} &15.28&0.265&9.10&&25.61&0.250&12.29 \\
         Stable Inpainting \cite{rombach2022high}&12.57&0.264&7.07&&18.13&0.246&8.50 \\
         \midrule
         SmartBrush (Ours) &\textbf{9.71}&\textbf{0.266}&\textbf{6.00}&&\textbf{13.22}&\textbf{0.252}&\textbf{8.05}\\
    \bottomrule
    \end{tabular}
    \label{tab:box_mask}
\end{table*}

\begin{table*}[]
    \centering
     \caption{Text-guided object inpainting with object layout mask.}
    \begin{tabular}{lcccp{3em}ccc}
    \toprule
    & \multicolumn{3}{c}{OpenImages} && \multicolumn{3}{c}{MSCOCO}\\
     &Local FID $\downarrow$ & CLIP Score $\uparrow$ &FID $\downarrow$ &&Local FID $\downarrow$ & CLIP Score $\uparrow$& FID $\downarrow$\\ \midrule
         Blended Diffusion \cite{avrahami2022blended}& 21.93&0.261&9.72&&26.25&0.244&8.16 \\
         GLIDE \cite{nichol2021glide} &21.09&0.250&9.03&&24.25&0.235&6.98\\ 
         Stable Diffusion \cite{rombach2022high}& 12.27&\textbf{0.263}&6.90&&17.16&0.246&7.78\\ 
         Stable Inpainting \cite{rombach2022high}& 10.98&0.261&5.84&&15.16&0.243&6.54\\
         \midrule
         SmartBrush (Ours) &\textbf{7.82}&\textbf{0.263}&\textbf{4.70}&&\textbf{9.80}&\textbf{0.249}&\textbf{5.76}\\
    \bottomrule
    \end{tabular}
    \label{tab:object_mask}
\end{table*}

\begin{figure}
    \centering
    \includegraphics[height=3.5cm, width=6cm]{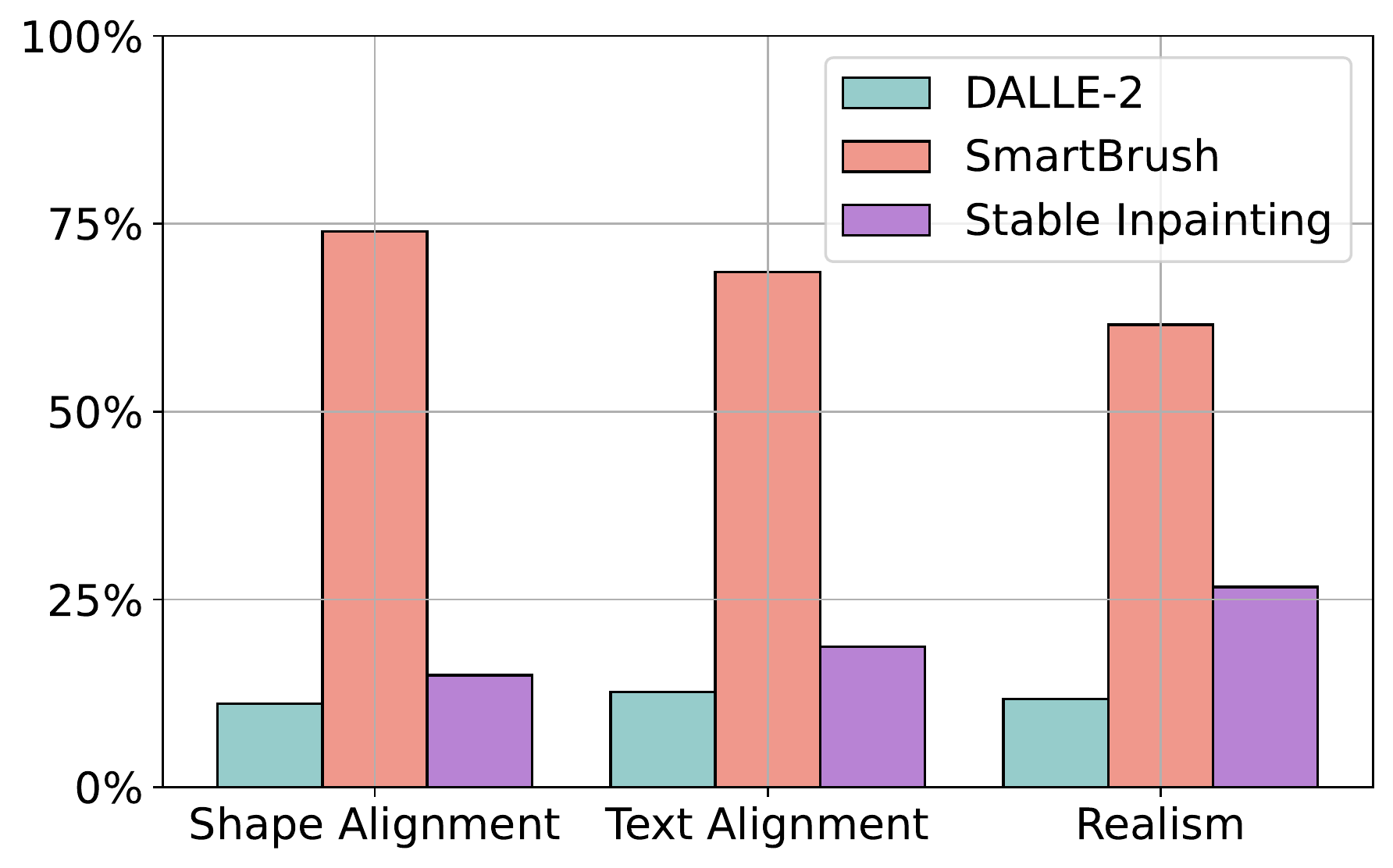}
    \caption{We ask users to choose the generation that best aligns with the mask and input text, and looks most realistic. Our method SmartBrush outperforms the baselines by a large margin.}
    \label{fig:user_study}
\end{figure}

\begin{figure}
\def\fsh{\small}
    \centering
    \def\qualwidth{1.2cm}
    \def\fsh{\footnotesize}
    \def\fsc{\footnotesize}
    \setlength{\tabcolsep}{2pt}
   \begin{tabular}{cccccc}
       \fsh{Input} & \fsh{Mask 0} & \fsh{Mask1} &\fsh{Mask2} & \fsh{Mask3}  & \fsh{Mask4}\\
       
       \frame{\includegraphics[width=\qualwidth]{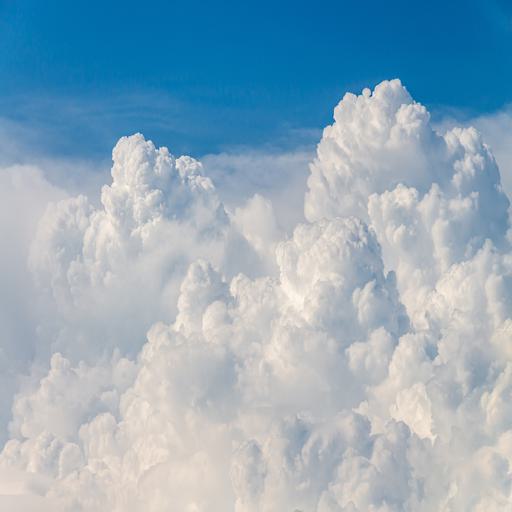}} &
        \includegraphics[width=\qualwidth]{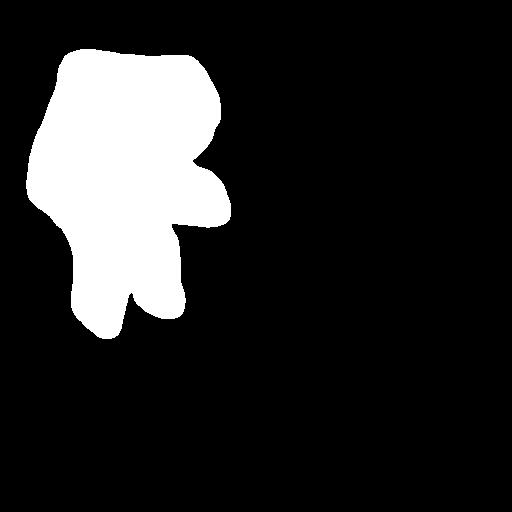} &
        \includegraphics[width=\qualwidth]{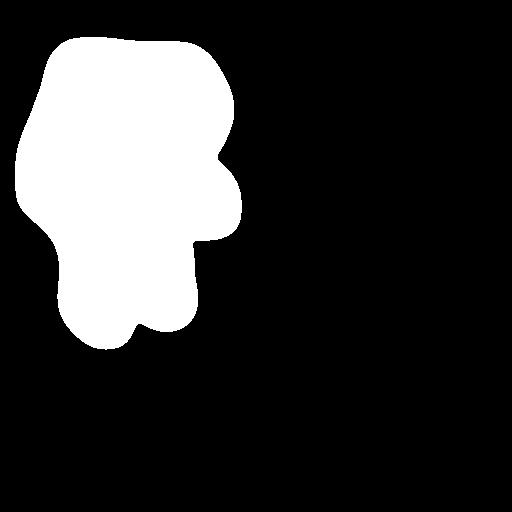} &
        \includegraphics[width=\qualwidth]{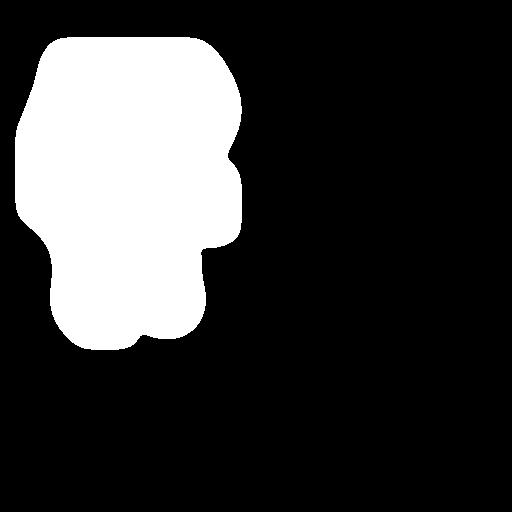} &
        \includegraphics[width=\qualwidth]{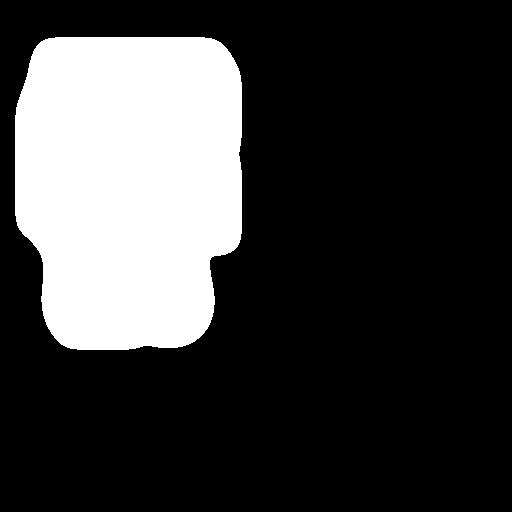} &
        \includegraphics[width=\qualwidth]{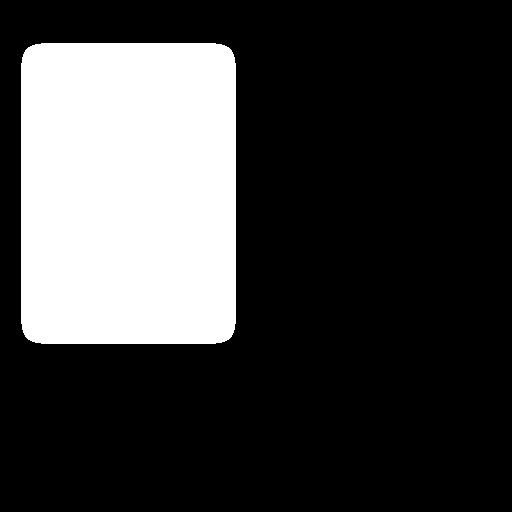} \\
        SDiffusion& 
       \frame{\includegraphics[width=\qualwidth]{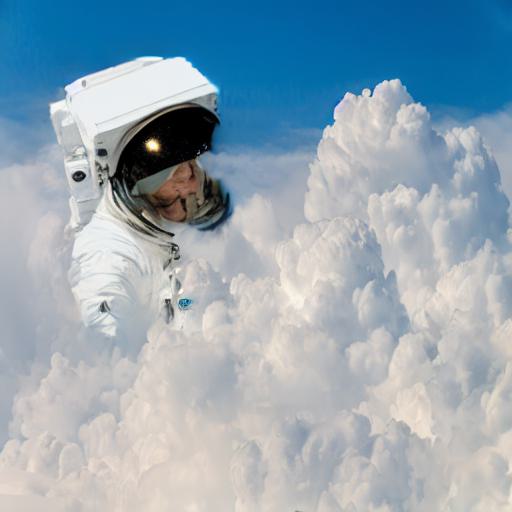}} & 
        \frame{\includegraphics[width=\qualwidth]{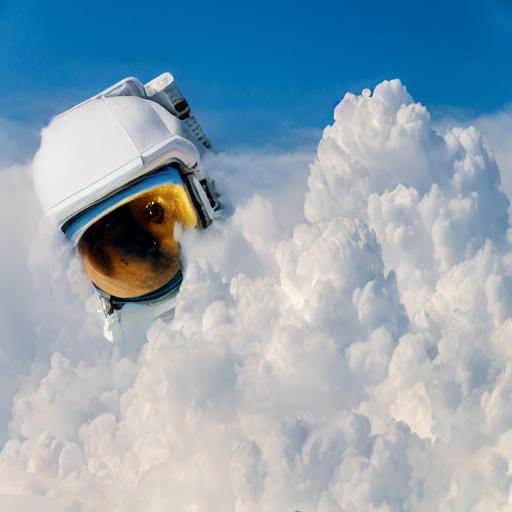}} & \frame{\includegraphics[width=\qualwidth]{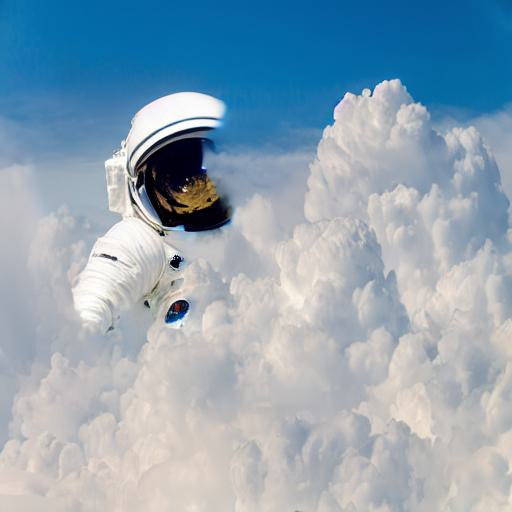}} & \frame{\includegraphics[width=\qualwidth]{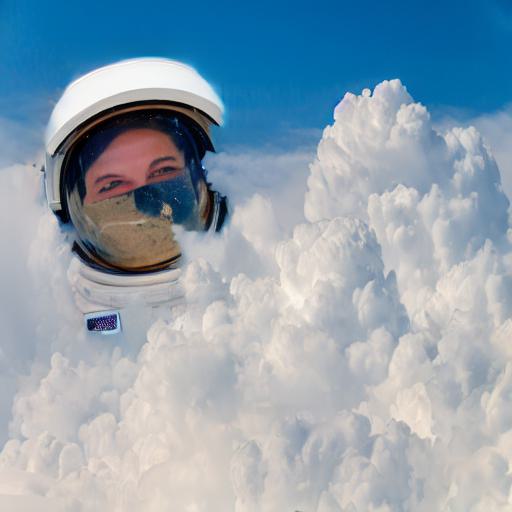}} &
        \frame{\includegraphics[width=\qualwidth]{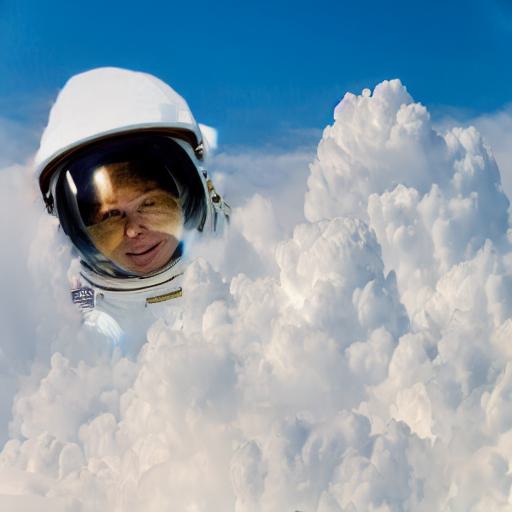}} \\
        SInpainting &
        \frame{\includegraphics[width=\qualwidth]{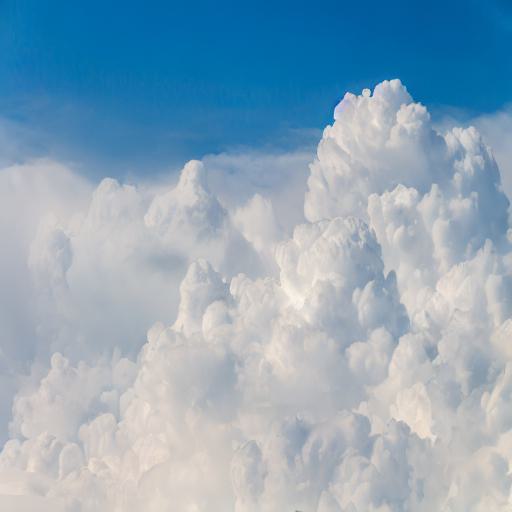}} &
        \frame{\includegraphics[width=\qualwidth]{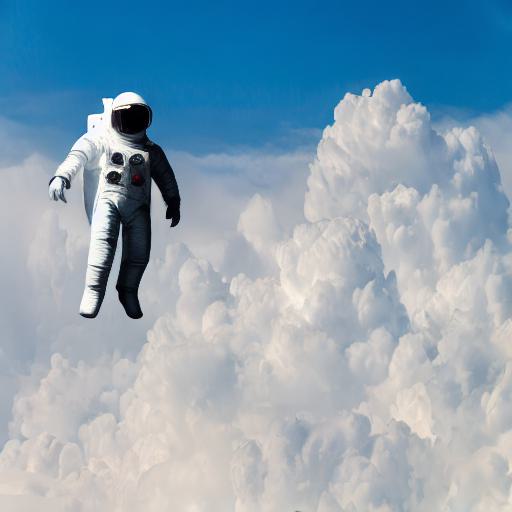}} &
        \frame{\includegraphics[width=\qualwidth]{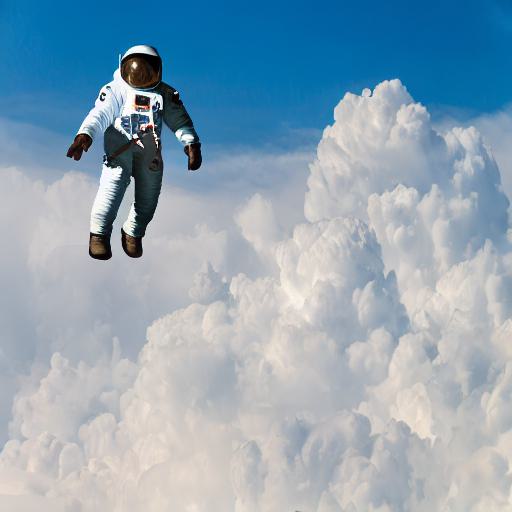}} &
        \frame{\includegraphics[width=\qualwidth]{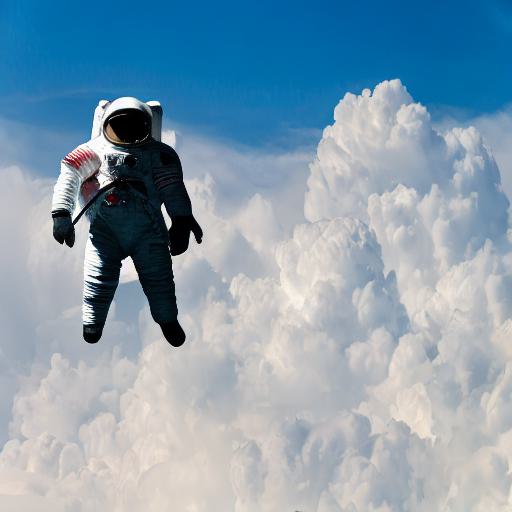}}  &
        \frame{\includegraphics[width=\qualwidth]{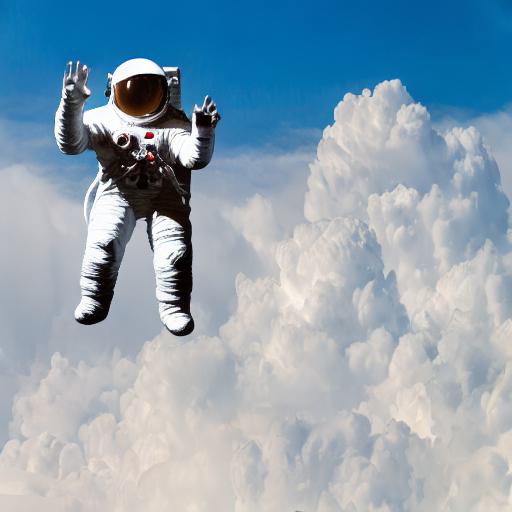}} 
        \\
        Ours & \frame{\includegraphics[width=\qualwidth]{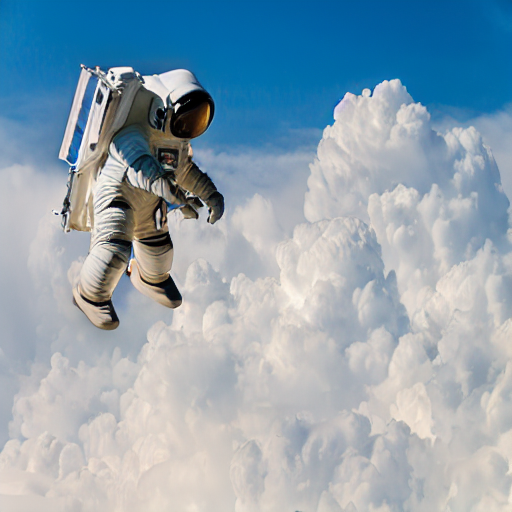}} 
        &
        \frame{\includegraphics[width=\qualwidth]{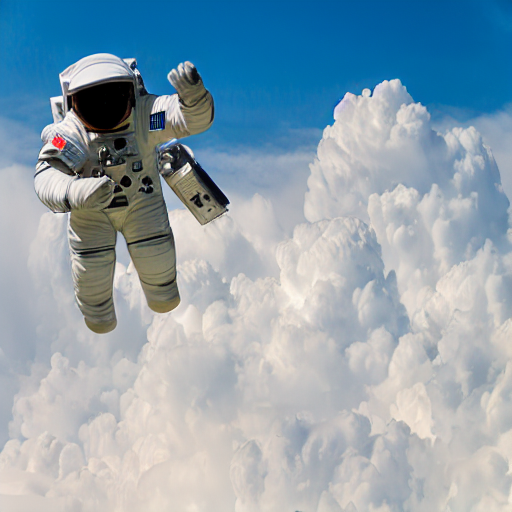}} 
        &
        \frame{\includegraphics[width=\qualwidth]{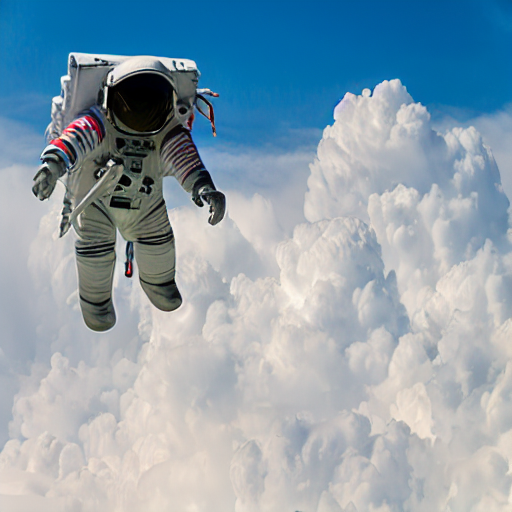}} 
        &\frame{\includegraphics[width=\qualwidth]{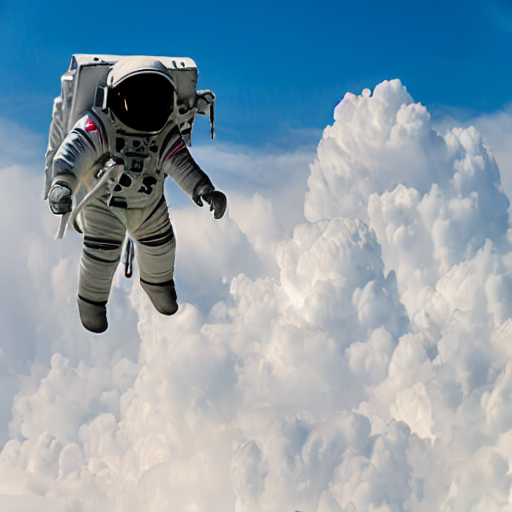}} 
        &\frame{\includegraphics[width=\qualwidth]{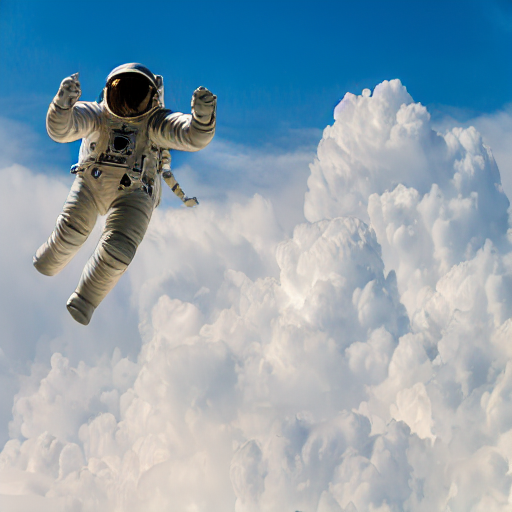}} \\
       
   \end{tabular}
    \caption{Mask precision control samples with prompt ``\emph{astronaut}''. 
    As we increase the mask type, our method give more freedom to the model and the outputs gradually become different from the input object shape mask.}
    \label{fig:mask_precision}
\end{figure}

\setlength{\tabcolsep}{1pt}
\begin{figure*}
    \centering
    \def\fsh{\small}
    \def\fsc{\small}
    \def\qualwidth{3.2cm}
    \begin{tabular}{cccccc}
    &\fsc{Input} & \fsc{Mask} & \fsc{DALLE-2} & \fsc{SmartBrush w/o bg pres} & \fsc{SmartBrush} \\
        \raisebox{4.5em}{\rotatebox[origin=c]{90}{\fsh{car}}}&\frame{\includegraphics[width=\qualwidth]{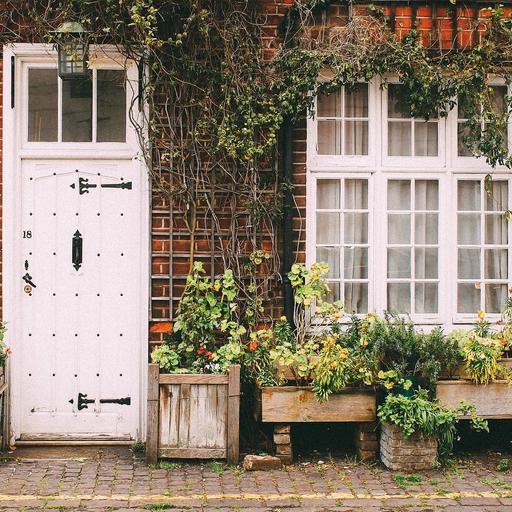}} & \frame{\includegraphics[width=\qualwidth]{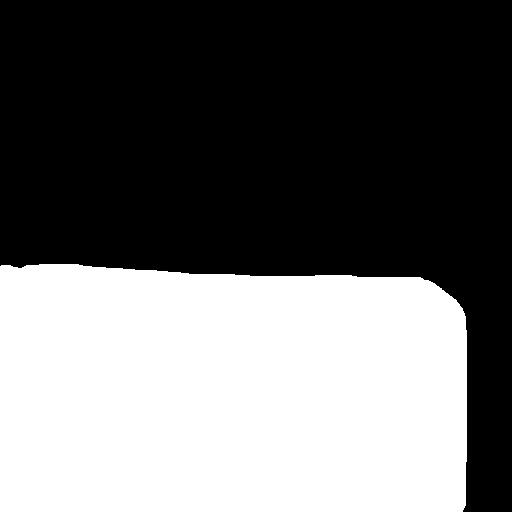}}&
        \frame{\includegraphics[width=\qualwidth]{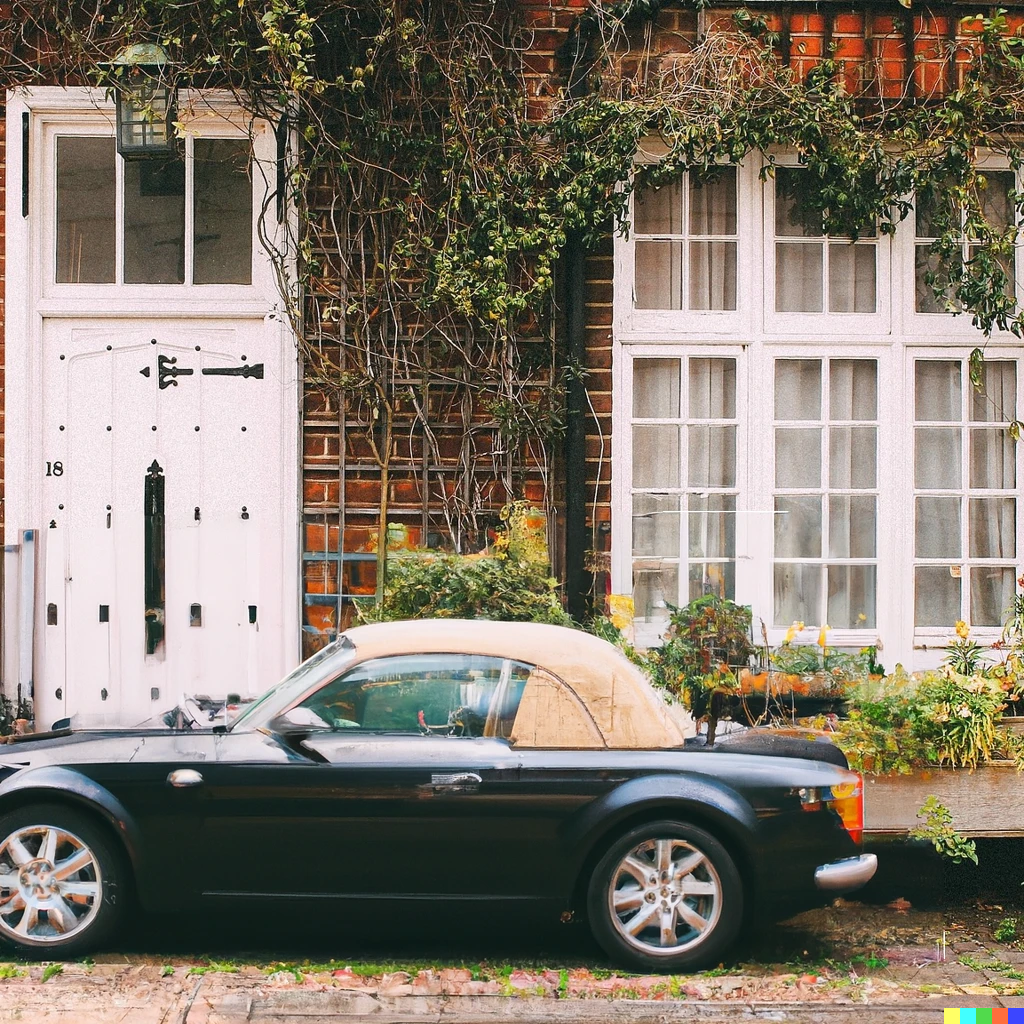}}
        &
        \frame{\includegraphics[width=\qualwidth]{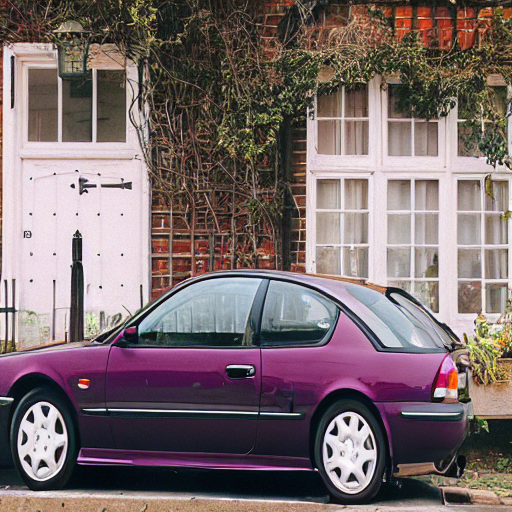}}
        &\frame{\includegraphics[width=\qualwidth]{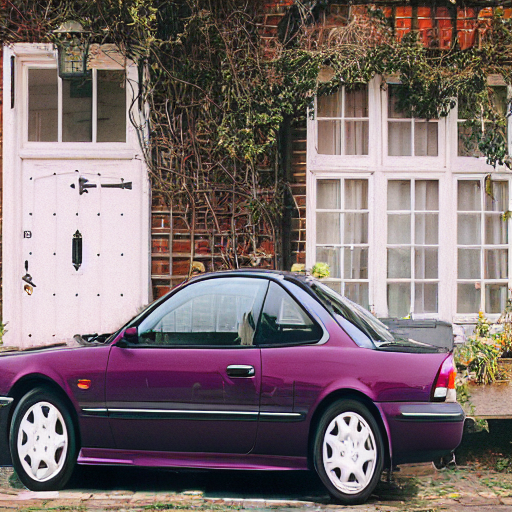}}\\
         \raisebox{4.5em}{\rotatebox[origin=c]{90}{\fsh{British shorthair cat}}}&\frame{\includegraphics[width=\qualwidth]{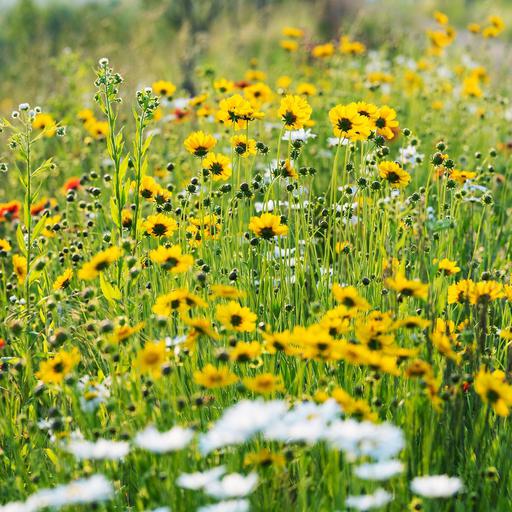}} & \frame{\includegraphics[width=\qualwidth]{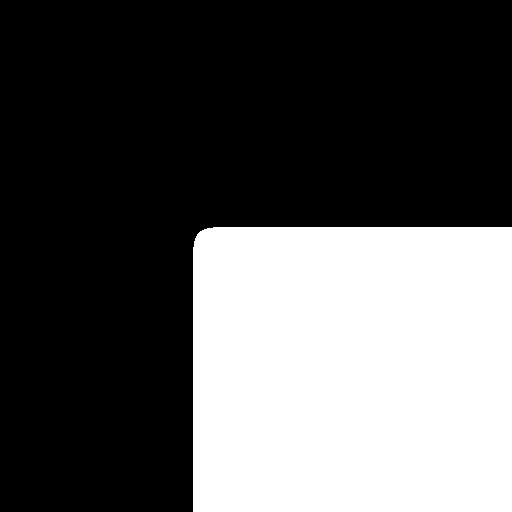}}&
        \frame{\includegraphics[width=\qualwidth]{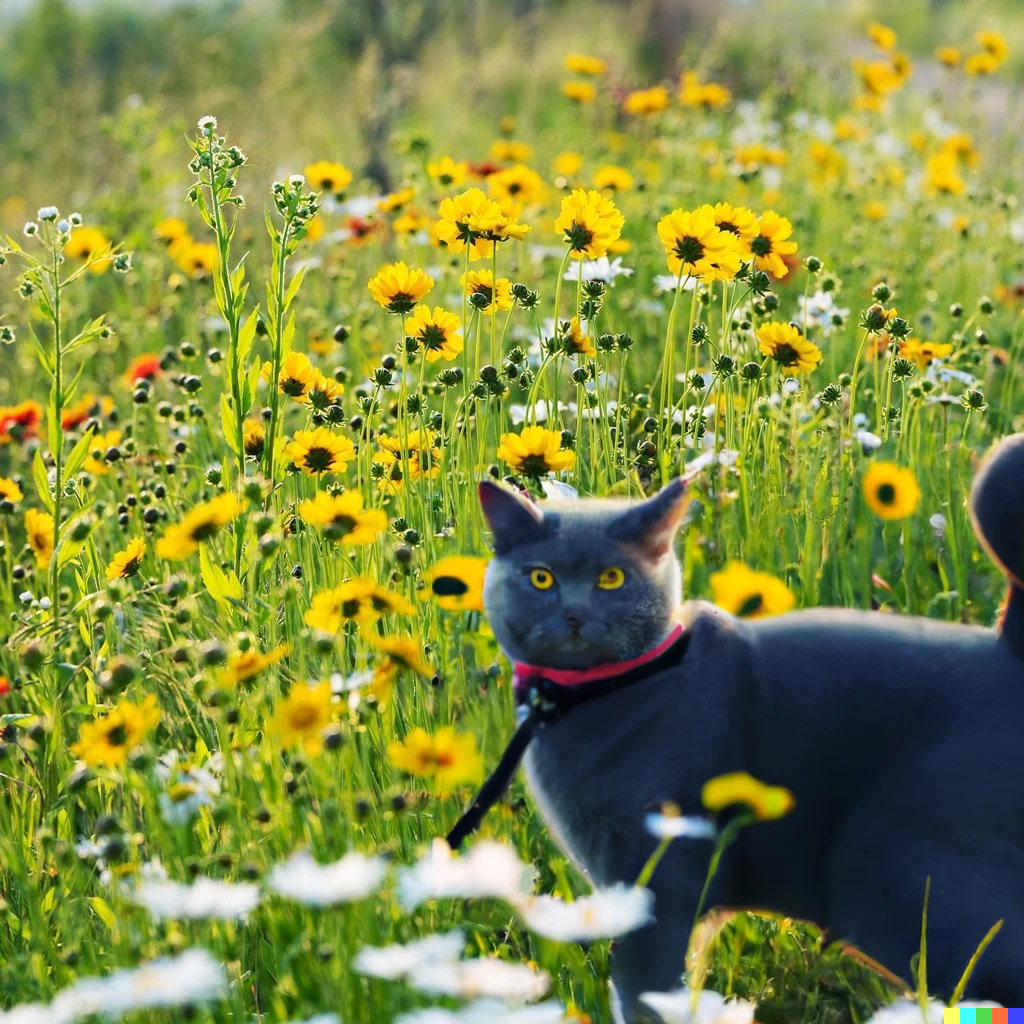}}
        &
        \frame{\includegraphics[width=\qualwidth]{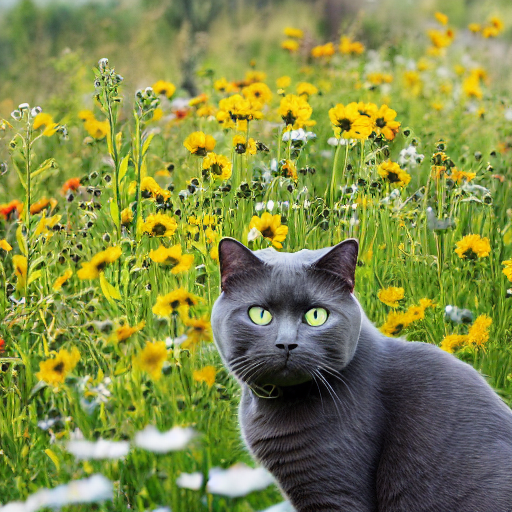}}
        &\frame{\includegraphics[width=\qualwidth]{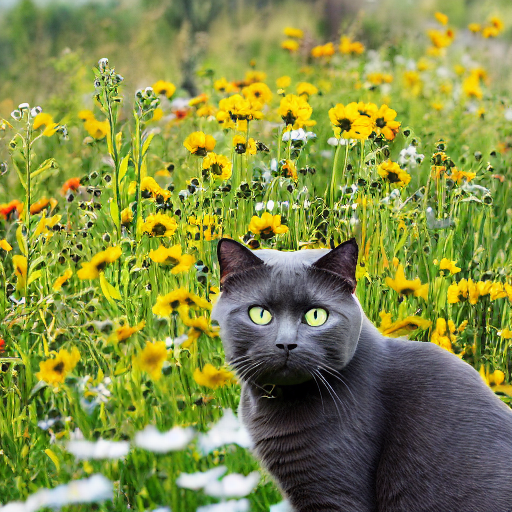}}\\
         \raisebox{4.5em}{\rotatebox[origin=c]{90}{\fsh{teddy bear}}}&\frame{\includegraphics[width=\qualwidth]{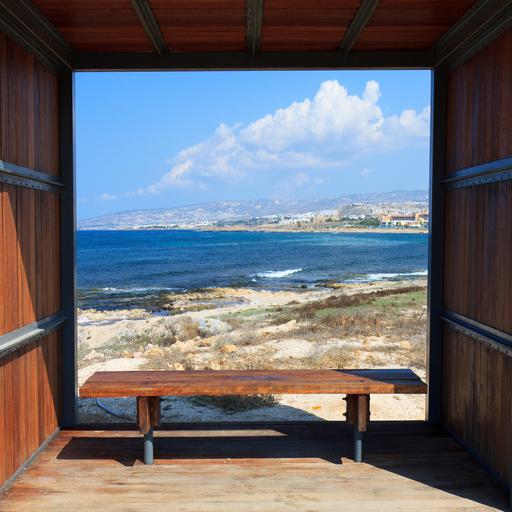}} & \frame{\includegraphics[width=\qualwidth]{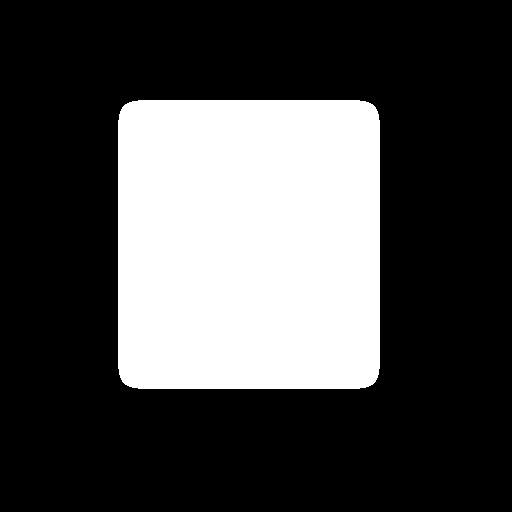}}&
        \frame{\includegraphics[width=\qualwidth]{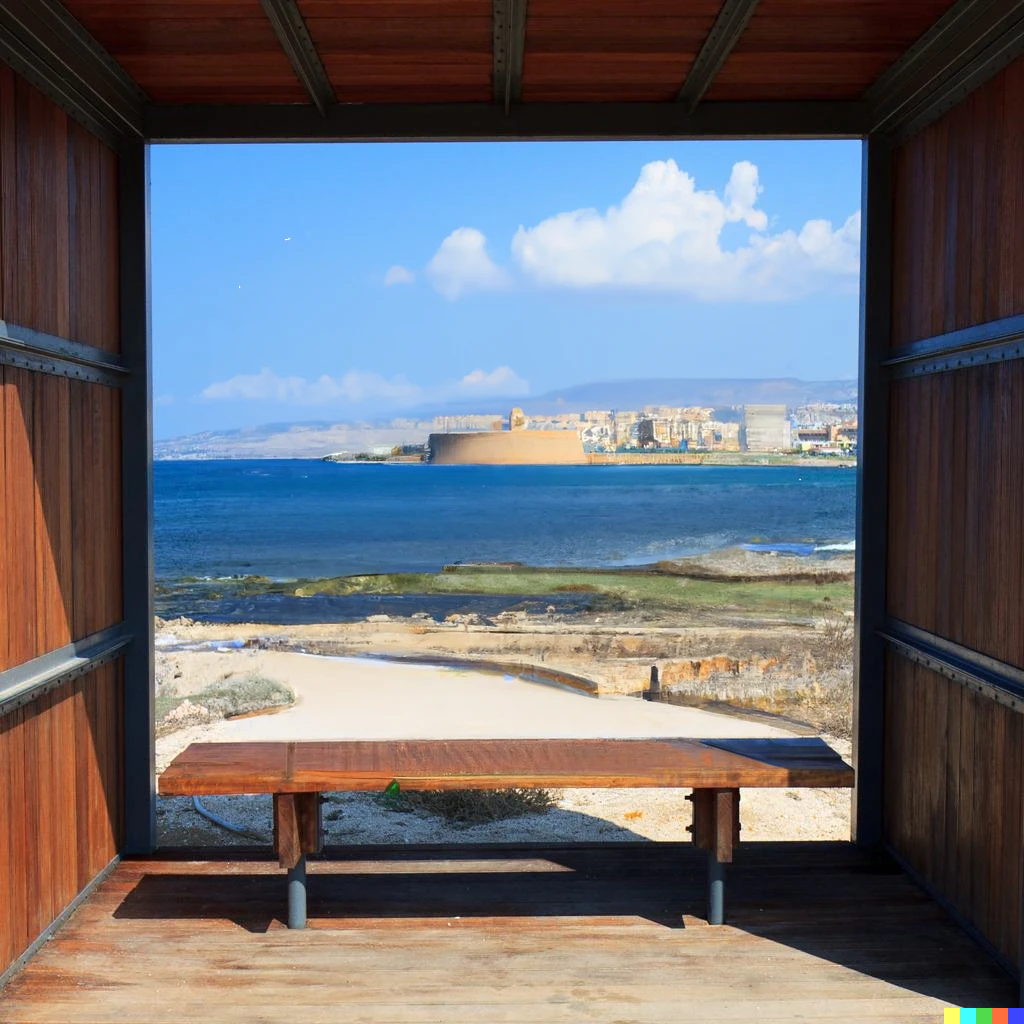}}
        &
        \frame{\includegraphics[width=\qualwidth]{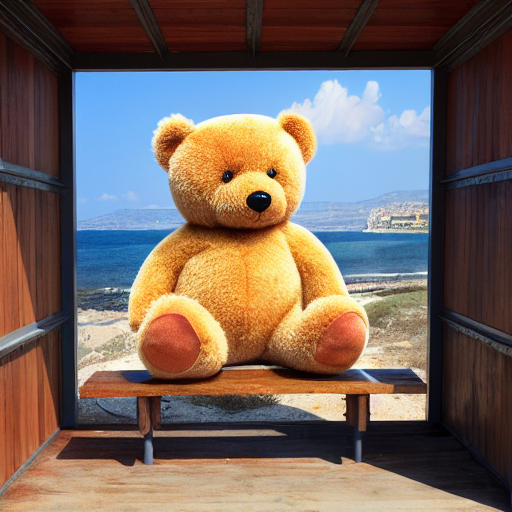}}
        &\frame{\includegraphics[width=\qualwidth]{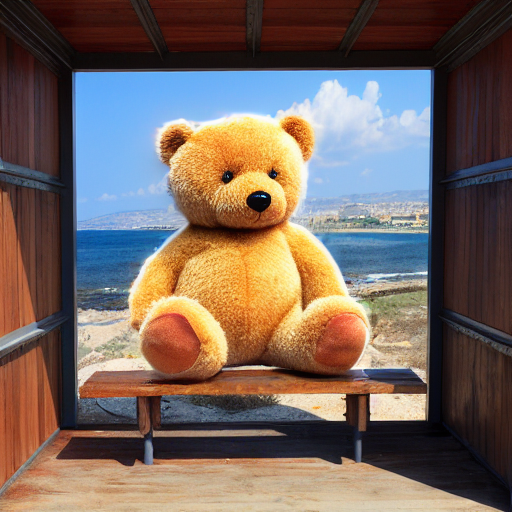}}\\
    \end{tabular}
    \caption{Comparison of background preservation in inpainting. We only compare with DALLE-2 here for better visualization, and more baseline results are provided in the supplementary. We observe that DALLE-2 and SmartBrush w/o background preservation change the background surrounding the generated object, \eg, the door bell, clouds behind the mountain, flowers behind the cat, and landscape behind the teddy bear. By contrast, our SmartBrush better preserves the background pixels. 
    }
    \label{fig:box_generation}
\end{figure*}

\begin{figure}
   \def\fsh{\small}
    \centering
   \begin{tabular}{c}
   \fsh{\hspace{-0.3cm}Input \hspace{0.45cm}0 \hspace{0.45cm}10 \hspace{0.45cm}20 \hspace{0.45cm}30 \hspace{0.45cm}40 \hspace{0.45cm}50} \\ 
         \includegraphics[scale=0.42]{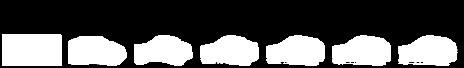}  \\
         \includegraphics[scale=0.42]{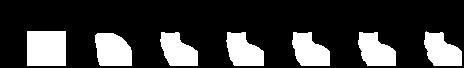}  \\
         \includegraphics[scale=0.42]{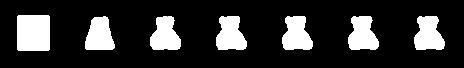}  \\
   \end{tabular}
    \caption{Predicted object masks corresponding to examples from \cref{fig:box_generation}. The numbers denote the sampling time steps. The mask prediction becomes sharper after around 10 steps.}
    \label{fig:box_prediction}
\end{figure}

\textbf{Baselines}
We choose the state-of-the-art image inpainting methods as our baselines, \ie, Blended Diffusion~\cite{avrahami2022blended}, GLIDE~\cite{nichol2021glide}, Stable Diffusion~\cite{rombach2022high}, and Stable Inpainting~\cite{rombach2022high}. We also compare with DALLE-2~\cite{ramesh2022hierarchical} on limited images since its model is not open source yet. Stable Diffusion, Stable Inpainting, and our SmartBrush support image generation on the size of 512$\times$512. Since Blended Diffusion and GLIDE only support images size of 256$\times$256, we resize all results to 256$\times$256 for fair comparison. 

\textbf{Testing Datasets}
We evaluate our model on two popular segmentation datasets, \ie, OpenImages~\cite{schuhmann2021laion} and MSCOCO~\cite{lin2014microsoft}. We sample 2 masks for each image in the testing dataset of MSCOCO, so the number of testing images is 9311. As for OpenImages, we sample images with resolution higher than 512 and use one mask for each image. Then, the number of testing images is 13400. The input prompts are directly from segmentation class labels.

\textbf{Evaluation Metrics}
We first measure the image quality by Frechet Inception Distance (FID)~\cite{Seitzer2020FID}. Since our main task is object generation in the masked region, the global FID cannot well reflect the generation quality since the masked region may occupy a small part of the image. Therefore, we crop the images according to the bounding box of the mask and measure FID on the local regions, which is referred to as ``Local FID''. To measure the alignment between text and generated content, we adopt the CLIP score~\cite{radford2021learning}.


\subsection{Text and Shape Guided Inpainting}
The proposed SmartBrush can inpaint not only objects but also generic scene like sunset sky by following the text and shape guidance.
For object inpainting, we consider two common use cases: 1) accurate object masks and 2) bounding box masks. The former expects the generated object to follow the given mask shape, while the latter does not constrain the shape of generated objects as long as they are inside of the box. Corresponding quantitative results are listed in \cref{tab:box_mask,tab:object_mask}.

As a strong baseline, Stable Inpainting presents lower CLIP scores than ours, which suggests that random masking is not an optimal training strategy for text-guided inpainting. The Blended Diffusion achieves a relatively high CLIP score but lags far behind in FID since the CLIP model focus on the global content instead of local objects. 
By contrast, our SmartBrush achieves the best performance in both tasks on all metrics, which demonstrates the effectiveness of our proposed training strategy with text and shape guidance.

\cref{fig:object_generation} visualizes inpainting examples from the baselines and our SmartBrush. In general, we can generate high-quality objects/scenes well following both the mask shape and text, no matter short words or long sentences. By contrast, all baselines failed following the mask shape. 
Even, Blended Diffusion and GLIDE cannot generate decent objects given these local text descriptions.
Stable Diffusion, Stable Inpainting, and DALLE-2 could be better but with high chance of misunderstanding the text caused by text misalignment.

Besides object inpainting, our SmartBrush also supports scene inpainting as illustrated by the last two rows in \cref{fig:object_generation}. More examples can be found in the supplementary. Still, as compared to our SmartBrush, it is difficult for existing inpainting models to follow the mask shape.

We also conduct user studies through Amazon Mechanical Turk. Over 300 workers were asked 1) which result follows the object mask best, 2) which result follows the input text description best, and 3) which result looks most natural/realistic. The survey result is shown in \cref{fig:user_study}, where more then 50\% users vote our results as the best on each question.



\subsection{Mask Precision Control}
In the real world, users will not always provide the precise mask of the object they want to inpaint. We may encounter a coarse mask, so SmartBrush accepts the control of how closely the inpainted object is to the given mask. \cref{fig:mask_precision} shows the results with different types of masks, which follow the blurring rule during training, \ie, applying Gaussian blur iteratively to obtain masks from fine to coarse. The Stable Diffusion results are not affected by mask types since it is not trained that way. The results of Stable Inpainting only change the object size with the mask size but do not follow the mask shape. 
By contrast, ours strictly follow the mask shape when providing a finer mask, while roughly following the mask if given a coarser mask. For extremely, given a box-like mask (the last column), we allow the generation to happen anywhere inside the box.

\subsection{Background Preservation}
To inpaint an object, especially when giving a box-like mask,  it is important to preserve the background since the inpainted object will only partially occupy the mask area.  
\cref{fig:box_generation} compares different methods in background preservation when giving box-like masks. Without any background preservation regularization, DALLE-2 generates objects inside the mask and changes the non-object pixels inside the mask. Our SmartBrush, with object mask prediction (shown in \cref{fig:box_prediction}), could much better preserve the background by utilizing the predicted mask during sampling.




\section{Conclusion, Limitation, and Future Work}
Existing text and shape guided image inpainting models face three typical challenges: mask misalignment, text misalignment, and background preservation. In this paper, we propose a novel training method that utilizes the text and shape guidance from the segmentation dataset to address the text misalignment problem. Then we further propose to create different levels of masks (from fine to coarse) to allow precision control of the generation. Finally, we propose an additional training loss function to encourage the model to make object predictions from the input box mask. Then we can utilize the predicted mask to avoid unnecessary changes inside the mask. The quantitative and qualitative results demonstrate the superiority of our method.

The main limitation of our method is the large shadow case, where the shadow of the object exceeds the object mask, \eg, the shadow of a person can be very long in the morning while the bounding box usually fails to cover the whole shadow. Our method may not be able to generate such long shadow since the coarsest mask is the object bounding box. We will explore it in the near future. 

{\small
\bibliographystyle{ieee_fullname}
\bibliography{egbib}
}

\end{document}